\documentclass[acmsmall,table]{acmart}
\acmSubmissionID{280}

\setcopyright{cc}
\setcctype{by}
\acmJournal{PACMMOD}
\acmYear{2025} \acmVolume{3} \acmNumber{6 (SIGMOD)} \acmArticle{377} \acmMonth{12} \acmPrice{}\acmDOI{10.1145/3769842}
\received{April 2025}
\received[revised]{July 2025}
\received[accepted]{August 2025} 

\usepackage{amsmath,amsfonts}
\usepackage{placeins}
\usepackage{algorithm}
\usepackage[noend]{algpseudocode}
\usepackage{xcolor}
\usepackage{graphicx}
\usepackage{makecell}
\usepackage{arydshln}
\usepackage{subcaption}
\usepackage{textcomp}
\usepackage{xspace}
\usepackage{colortbl}
\usepackage{pifont}
\usepackage{booktabs}
\usepackage{multirow}
\usepackage{hyperref}
\usepackage{url}
\usepackage{bm}
\usepackage[framemethod=TikZ,
innerleftmargin=0.1\baselineskip]{mdframed}
\usepackage{censor}
\newmdenv[linecolor=white,backgroundcolor=mygray]{myframe}
 \definecolor{mygray}{gray}{0.95}
\newcommand*\colourcheck[1]{%
  \expandafter\newcommand\csname #1check\endcsname{\textcolor{#1}{\ding{52}}}%
}
\newcommand*\colourcross[1]{%
  \expandafter\newcommand\csname #1cross\endcsname{\textcolor{#1}{\ding{54}}}%
}
\newcommand{\bemph}[1]{\textbf{\textit{#1}}}

\newcommand{\parab}[1]            

\colourcheck{orange}
\colourcross{red}
\colourcheck{teal}
\newcommand{\algo}{\textsc{WaveStitch}\xspace}
\newcommand{\algoar}{\textsc{WaveStitchAR}\xspace}

\definecolor{colblue}{rgb}{0.42353,0.55686,0.74902}

\definecolor{colora}{rgb}{0.84314,0.60784,0}

\definecolor{colorgr}{rgb}{0.18431,0.59216,0.30588}

\definecolor{darkred}{rgb}{0.55, 0.0, 0.0}

\definecolor{darkseagreen}{rgb}{0.24, 0.70, 0.44}
\definecolor{lightgray}{gray}{0.95}
\definecolor{olivebg}{rgb}{0.9, 0.92, 0.82}

\begin{document}
\title{WaveStitch: Flexible and Fast Conditional Time Series Generation with Diffusion Models}

\author{Aditya Shankar}
\affiliation{%
  \institution{Delft University of Technology}
  \city{Delft}
  \country{The Netherlands}
}
\email{a.shankar@tudelft.nl}

\author{Lydia Chen}
\orcid{}
\affiliation{%
  \institution{Delft University of Technology, Université de Neuchâtel}
  \city{Neuchâtel}
  \country{Switzerland}
}
\email{lydiaychen@ieee.org}

\author{Arie van Deursen}
\orcid{}
\affiliation{%
  \institution{Delft University of Technology}
  \city{Delft}
  \country{The Netherlands}
}
\email{arie.vandeursen@tudelft.nl}

\author{Rihan Hai}
\affiliation{%
  \institution{Delft University of Technology}
  \city{Delft}
  \country{The Netherlands}
}
\email{r.hai@tudelft.nl}



\begin{abstract}
Generating temporal data under {\color{black}conditions} is crucial for forecasting, imputation, and generative tasks. Such data often has {\color{black}metadata} and partially observed {\color{black} signals} that jointly influence the generated values. However, existing methods face three key limitations: {\color{black}(1) they condition on either the metadata or observed values, but rarely both together; (2) they adopt either training-time approaches that fail to generalise to unseen scenarios, or inference-time approaches that ignore metadata; and (3) they suffer from trade-offs between generation speed and temporal coherence across time windows, choosing either slow but coherent autoregressive methods or fast but incoherent parallel ones.} We propose {WaveStitch}, a novel diffusion-based method to overcome these hurdles through: {\color{black}(1) dual-sourced conditioning on both metadata and partially observed signals; (2) a hybrid training-inference architecture, incorporating metadata during training and observations at inference via gradient-based guidance; and (3) a novel pipeline-style paradigm that generates time windows in parallel while preserving coherence through an inference-time conditional loss and a stitching mechanism.} Across diverse datasets, WaveStitch demonstrates adaptability to arbitrary patterns of observed signals, achieving {\color{black}{1.81x}} lower mean-squared-error compared to the state-of-the-art, and generates data up to {166.48x} faster than autoregressive methods while maintaining coherence. Our code is available at: \url{https://github.com/adis98/WaveStitch}.
\end{abstract}

\ccsdesc[500]{Information systems~Data management systems}

\keywords{Time Series Generation, Conditional Generation, Diffusion Models}

\maketitle

\section{Introduction}
\label{sec:intro}

{\color{black}Time series data often suffer from missing or noisy values, or are subject to access restrictions due to privacy concerns, which are important challenges in data management. Although synthetic data generation shows promise in mitigating these issues~\cite{park2018data,shankar2024silofuse,liu2024controllable}}, most generative models are \textit{unconditional}, i.e., they cannot control the characteristics of the generated data. 
\begin{figure}[tb]
    \centering
    \includegraphics[width=\textwidth]{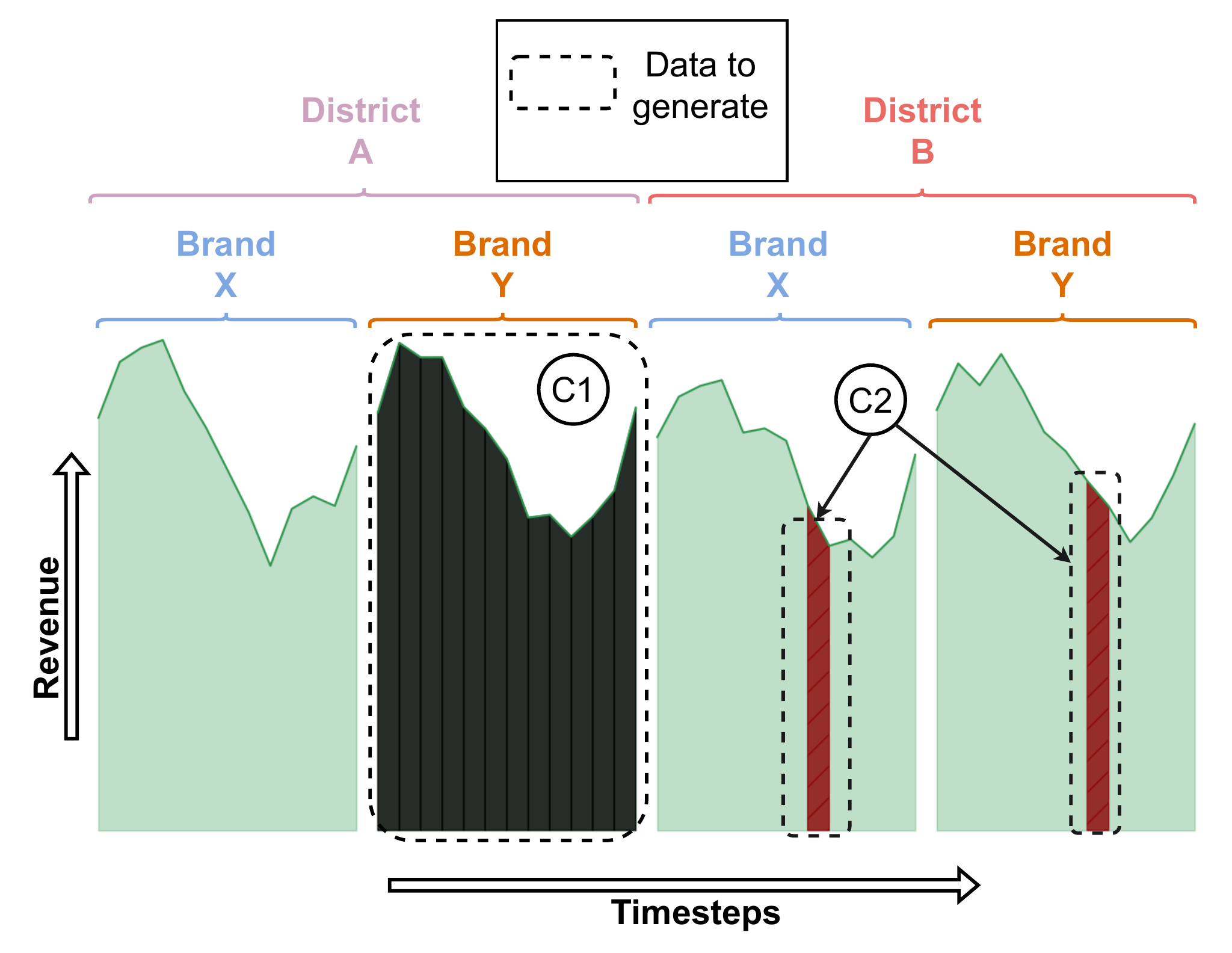}
    \caption{Time series (signal) of partially observed yearly revenues (in \textcolor{darkseagreen}{green}), grouped by metadata {District, Brand and Month}. We want to generate missing data under varying conditions, like C1: ({District} \textit{A} + {Brand} \textit{Y}), or C2: ({District \textit{B}} + {Month} \textit{August}).}
    \label{fig:arbitconst}
    \vspace{-0.47cm}
\end{figure}

\textit{Conditional} synthesis enables injecting user-defined constraints into the generation process, such as provided metadata or partially observed signals~\cite{narasimhan2024time,kollovieh2024predict,alcaraz2022diffusion}. {\color{black} For example, we can use it for data management tasks such as \textit{imputation}~\cite{imdiffusion}, \textit{forecasting}~\cite{faloutsos2018forecasting,qiu2024tfb}, and data cleaning~\cite{sigmoddatacleaning} by providing clean parts of the data as conditional observations for the model to regenerate or correct the rest.} For such tasks, forecasting is often treated as a special case of imputation, where missing values occur only at the tail-end of the time series ~\cite{alcaraz2022diffusion,kollovieh2024predict}. Going beyond imputation and forecasting, we can also have \textit{generative} tasks that are more loosely defined and are used for tasks like data augmentation or simulating new scenarios~\cite{liu2024controllable,sanghi2023synthetic,narasimhan2024time}. Unlike forecasting and imputation, generative tasks target producing samples that are consistent and plausible with high-level constraints such as metadata, rather than having an exact match with partially observed signals.

Consider \autoref{fig:arbitconst} showing a time series signal, \textit{Revenue}, accompanied by \color{black}{metadata}  \textit{District}, \textit{Brand}, and \textit{Month}. Suppose the data is stored by concatenating records from all district–brand combinations into a single {denormalised} table, with the row indices defining the timesteps. We can impose conditions by specifying the metadata. For example, a condition such as {District} \textit{A} + {Brand} \textit{Y} means "\textit{Generate revenue values for all timesteps with district A and brand Y.}" The model then generates values for all the timesteps matching the condition. However, this task is not straightforward due to the following challenges.


\textbf{Challenge 1.} 
Existing methods fail to condition on both the metadata (e.g., brand, district) \textit{and} the {\color{black}observed signals} in conjunction. \autoref{fig:arbitconst} shows that data can exhibit varying missingness patterns, such as contiguous blocks (C1), or fine-grained gaps (C2). Existing models condition only on the observed values and ignore the metadata~\cite{alcaraz2022diffusion,kollovieh2024predict}, or vice versa~\cite{narasimhan2024time}. {\color{black}Using only metadata can capture global trends, but would fail to align with the observed signals. Conversely, conditioning only on the observed signals may locally fill in gaps but overlook the broader context given by the metadata. The design challenge lies in using both sources to account for their differing influence in shaping global and local temporal patterns.}  

\textbf{Challenge 2.} 
The second challenge is deciding \textit{when} to incorporate conditional information, during \textit{training} or \textit{inference}. The choice is influenced by the differing characteristics of metadata and observed values. Metadata is static, and a model can learn its conditional influence during training. In contrast, signals may be observed at arbitrary timesteps (see C1 and C2 in \autoref{fig:arbitconst}), requiring flexibility at inference. However, existing works adopt a \textit{black-or-white} stance, conditioning exclusively during training \cite{narasimhan2024time, tashiro2021csdi, rasul21a, suh2024timeautodiff, alcaraz2022diffusion} or during inference \cite{kollovieh2024predict,lugmayr2022repaint}.

\begin{figure}
    \centering
    \begin{subfigure}{0.45\textwidth}
        \centering
        \includegraphics[width=\textwidth]{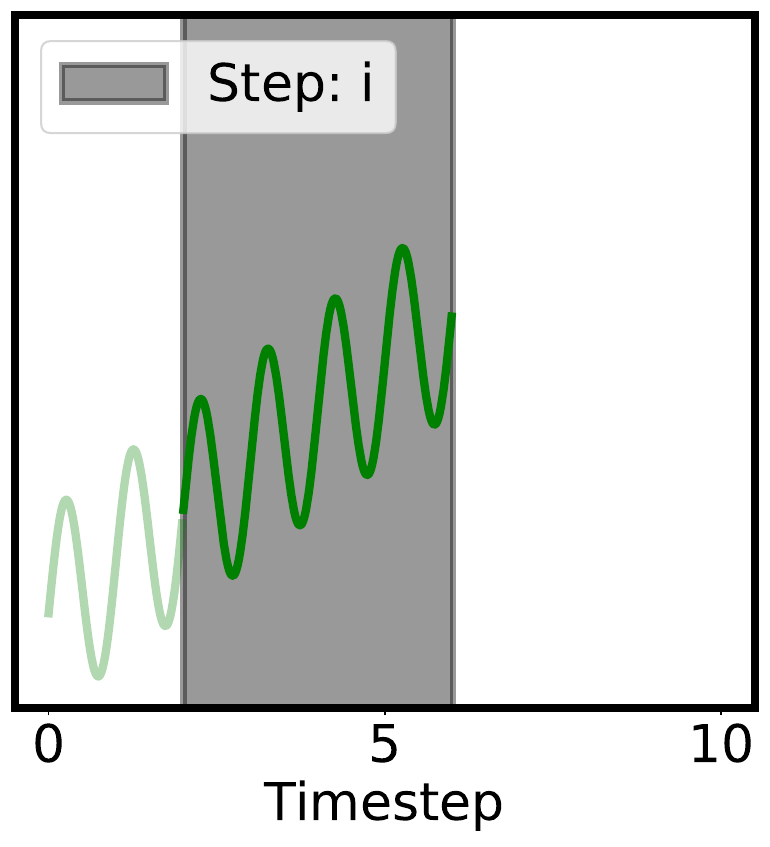}
        \caption{Signal generation Step $i$}
        \label{fig:enter-label}
    \end{subfigure}
    \begin{subfigure}{0.45\textwidth}
        \centering
        \includegraphics[width=\textwidth]{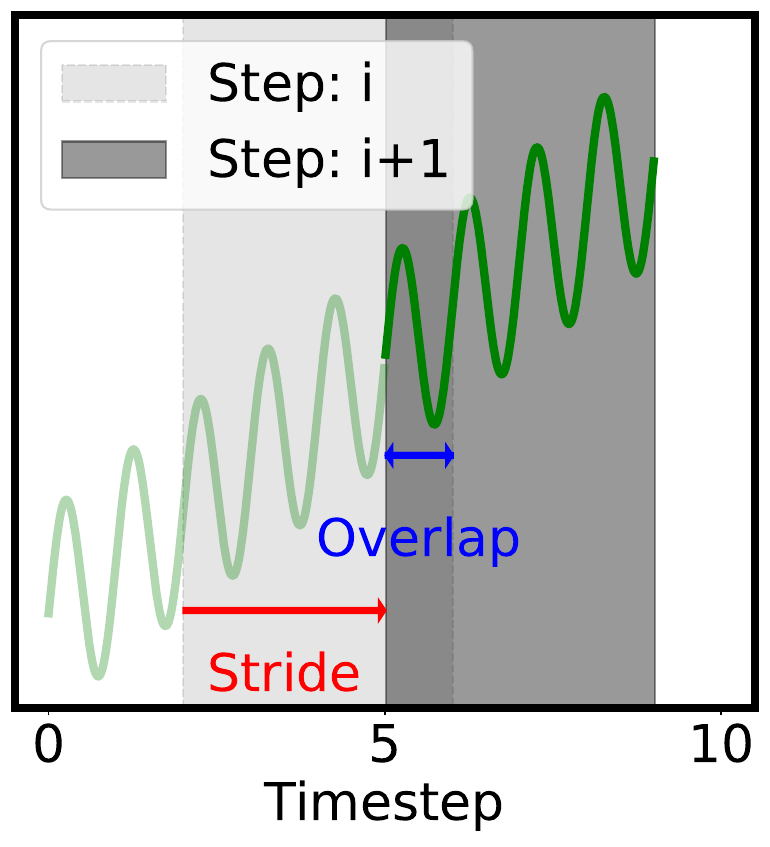}
        \caption{Signal generation Step $i+1$}
        \label{fig:enter-label}
    \end{subfigure}
    \caption{Autoregressive synthesis with sliding windows.}
    \label{fig:autoreg}
    \vspace{-0.3cm} 
\end{figure}

\textbf{Challenge 3.} Time series signals can be generated either \textit{autoregressively} or in \textit{parallel}. Autoregression incurs high runtime costs, generating signals sequentially using a sliding window to propagate temporal dependencies across consecutive segments~\cite{rasul21a} (see \autoref{fig:autoreg}). In contrast, parallel methods generate multiple windows simultaneously~\cite{kollovieh2024predict,narasimhan2024time}, but ignore maintaining temporal coherence between them. The challenge lies in maintaining both efficiency \textit{and} coherence, a balance that current methods struggle to achieve.

We propose \algo to tackle these challenges using \emph{denoising diffusion probabilistic models} (DDPMs)~\cite{ho2020denoising}. These models have demonstrated better performance over alternatives such as generative adversarial networks (GANs)~\cite{goodfellow2020generative} and Variational Autoencoders (VAEs)~\cite{kingma2019introduction} across various modalities~\cite{dhariwal2021diffusion,narasimhan2024time,kollovieh2024predict,kotelnikov2023tabddpm}, resulting in their adoption for data management tasks such as tabular and time series generation~\cite{shankar2024silofuse,liu2024controllable,imdiffusion}.

{\color{black}
Our key novelty is a unified framework to handle all three challenges. \algo \textit{dual-sources} its conditioning to use both metadata \textit{and} observed signals. This combination requires \textit{hybridising} training and inference. We first train a base generative model that conditions only on metadata. We inject the observed signals directly \textit{at inference}, guiding the sampling trajectory using gradients from a \textit{conditional loss}. This loss incorporates a novel \textit{stitching} mechanism to enforce coherence across overlapping time windows, thereby restricting the generation process to remain close to the realistic space. The approach is similar to \textit{gradient inversion}~\cite{jeon2021gradient}, where gradients update the \textit{sample} to match a given objective while keeping the model frozen. Moreover, the conditional loss is easily parallelised, enabling fast synthesis. This innovation allows us to break away from prior approaches and introduce \textbf{a new paradigm for time series generation} through \textit{pipeline parallelism}. Multiple time windows are generated simultaneously with dependencies propagating across overlaps, maintaining high efficiency and temporal coherence.}

To reiterate, we summarise our key contributions as follows:  
\\
{\color{black}\textbf{1. Dual-sourced Conditioning:}
\algo combines two complementary sources of information: metadata for capturing global context (Sec.~\ref{subsec:tr}), and observed signals (Sec.~\ref{subsec:inf}) for providing fine-grained local cues to refine generation with high precision. \\
\textbf{2. Hybrid Architecture:} \algo combines the strengths of training and inference-time strategies. It learns the conditional influence of the (static) metadata during training (Sec.~\ref{subsec:tr}), and dynamically refines samples at inference using any observed signal values via gradient-based guidance (Sec.~\ref{subsec:inf}). This approach enables the handling of arbitrary patterns of observed signals while preserving the global structure dictated by the metadata. 
\\
\textbf{3. Coherent Pipelined-Parallel Generation:}
To improve efficiency, we generate consecutive time windows in a \textit{pipelined-parallel} fashion, incorporating a novel \textit{stitching mechanism} to reconcile overlaps at inference. Dependency information propagates across time windows while they are generated in parallel. This process resembles a pipeline, ensuring smooth and coherent transitions without compromising speed (Secs.~\ref{subsec:inf},~\ref{sec:para}).}
\\
\textbf{4. Evaluations on Diverse Tasks:}
We evaluate \algo on point-wise and synthetic data quality metrics, demonstrating its flexibility across tasks and achieving up to {\color{black}\textbf{1.81x}} lower Mean-Squared-Error (MSE) compared to the state-of-the-art (SOTA) baselines. Results show that \algo's parallelism enables significantly faster generation than autoregressive synthesis (\textbf{166.48x}) while maintaining generation quality. Additional ablations assess the impact of the conditioning strategy, stitch loss formulations, and random imputation tasks with varying degrees of missingness (Sec.~\ref{sec:eva}).

\section{Background and Problem Definition}
We cover the foundations of diffusion models for time series, their conditional extensions, and formally define the problem while briefly highlighting the research gaps. \autoref{tab:notation} summarises all the key symbols and notations used henceforth.

 \subsection{Diffusion Models for Time Series}
 \label{subsec:diffforts}
\begin{figure}
    \centering
    \includegraphics[width=\textwidth]{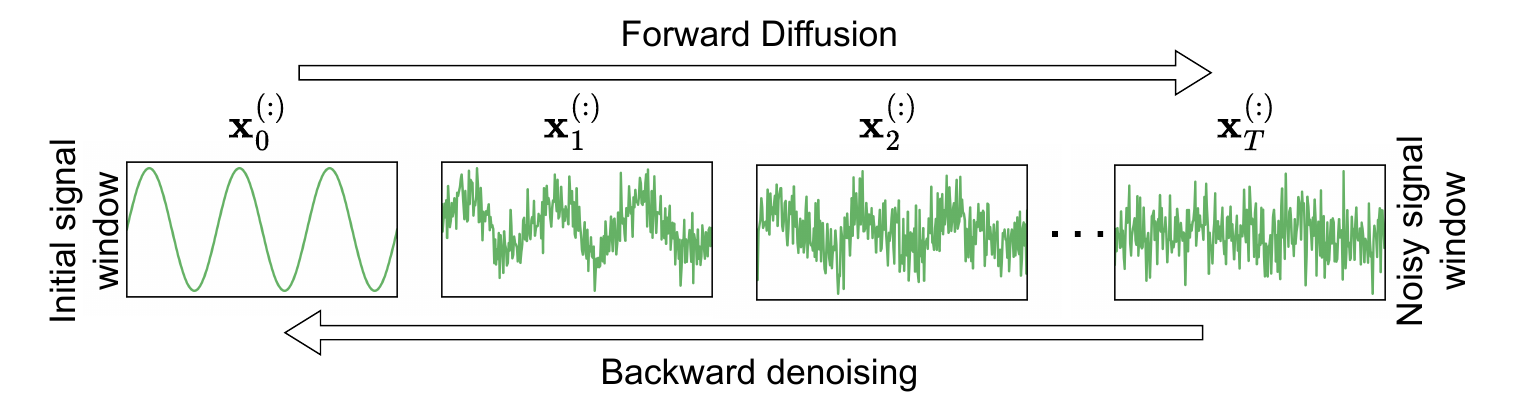}
    \caption{Diffusion process. $\mathbf{x}^{(:)}_0$ and $\mathbf{x}^{(:)}_T$ represent the clean and fully noised time window respectively. }
    \label{fig:fwdbckwd}
    \vspace{-10pt}
\end{figure}

 Denoising Diffusion Probabilistic Models (DDPMs) are a class of generative models used extensively for time series~\cite{alcaraz2022diffusion,kollovieh2024predict,yuan2024diffusion,narasimhan2024time}.
As shown in \autoref{fig:fwdbckwd}, DDPMs operate by gradually corrupting data with a \textit{forward} (noising) process, and then learning to reverse this through a backward \textit{denoising} process~\cite{ho2020denoising}. We forward noise a time window of $w$ timesteps starting from timestep $i$, as follows\footnote{There are two types of "steps" defined in this study. \textit{Diffusion} steps refer to the iterative noise addition or removal in the generative process, while \textit{timesteps} refer to the temporal axis in the time series data.}:

\begin{equation}
\label{eq:noising}
    \mathbf{x}^{(i:i+w-1)}_{t} = \sqrt{\alpha_t} \mathbf{x}^{(i:i+w-1)}_{t-1} + \sqrt{1 -\alpha_t} \epsilon_t,
\end{equation}
\noindent where \( \mathbf{x}^{(i:i+w-1)}_{t} \) is the multivariate time series signal after noising \( t\in\{0,1 \dots T\} \) steps, \( 1 - \alpha_t \) is the $t$-th noise variance, $(i:i+w-1)$ indicates the time slice\footnote{both $i$ and $i+w-1$ are included, yielding window size $w$}, and \( \epsilon_t \sim \mathcal{N}(0, I) \) is sampled white noise. With this notation, $\mathbf{x}^{(:)}_0$ and $\mathbf{x}^{(:)}_T$ represent a clean and fully noised sample, respectively. Noising up to any step without recursively applying eq.~\eqref{eq:noising} is done with a \emph{reparameterisation trick}~\cite{ho2020denoising}:

\begin{equation}
    \mathbf{x}^{(:)}_{t} = \sqrt{\bar{\alpha}_t} \mathbf{x}^{(:)}_{0} + \sqrt{1 - \bar{\alpha}_t} \epsilon,
    \label{eq:ffwdnoising}
\end{equation}
where \( \bar{\alpha}_t = \prod_{s=1}^t \alpha_s \) and \( \epsilon \sim \mathcal{N}(0, I) \). The reverse process iteratively denoises \( \mathbf{x}^{(:)}_{t} \) by estimating the noise at each step using a neural network \( f_{\theta}(\hat{\mathbf{x}}^{(:)}_{t}, t)\). This step is given by:

\begin{equation}
    \label{eq:denoiseone}
    \hat{\mathbf{x}}^{(:)}_{t-1} = \frac{1}{\sqrt{\alpha_t}} \left(\hat{\mathbf{x}}^{(:)}_{t} - \frac{1 - \alpha_t}{\sqrt{1 - \bar{\alpha}_t}} f_{\theta}(\hat{\mathbf{x}}^{(:)}_{t}, t) \right) + \sigma_t z,
\end{equation}
where $\hat{\mathbf{x}}^{(:)}_t$ and $\hat{\mathbf{x}}^{(:)}_{t-1}$ represent the signal estimates at step $t$ and $t-1$ respectively, \( \sigma_t \) is a function of $\alpha$ to ensure sample diversity, and \( z \sim \mathcal{N}(0, I) \)~\cite{ho2020denoising}. The denoiser trains by noising the signal up to a random step $t$ using eq.~\eqref{eq:ffwdnoising}, and then uses the MSE loss with the sampled noise for backpropagation, pushing the sample towards matching the ground truth, i.e., $\hat{x}_0^{(:)} \simeq x_0^{(:)}$~\cite{ho2020denoising}.

\subsection{Conditional Diffusion  for Time Series}
\label{subsec:conditionalddpm}

Standard DDPMs cannot generate samples adhering to conditions, as synthetic samples are generated freely from random noise in an unconstrained manner. In contrast, \emph{conditional} DDPMs constrain the denoising process using metadata or observed signals~\cite{alcaraz2022diffusion,narasimhan2024time,kollovieh2024predict}. These conditions can be incorporated at different times, either during \emph{training} or \emph{inference}.

\subsubsection{Training-time Conditioning} There are three main ways to condition a DDPM during training. The first way is to provide the conditions as additional inputs (e.g. metadata~\cite{narasimhan2024time}) during training, or through binary masks to indicate the timesteps with missing and observed signals ~\cite{alcaraz2022diffusion,tashiro2021csdi}. The second way is \textit{classifier-free guidance}~\cite{ho2021classifier,suh2024timeautodiff}, which trains both a conditional and an unconditional model, and interpolates between their outputs to control the degree of conditioning. The third way is \textit{classifier guidance}~\cite{liu2024controllable,dhariwal2021diffusion}, where an external classifier or controller guides the sampling of an unconditional denoiser~\cite{liu2024controllable}. The classifier functions as a feedback loop, adjusting the model's outputs by estimating the likelihood of meeting the given conditions at each denoising step.

\subsubsection{Inference-Time Conditioning} These approaches control sampling directly at inference by steering or guiding the outputs of an unconditional diffusion model. \textit{RePaint}~\cite{lugmayr2022repaint,imdiffusion}, adds a controlled amount of noise to the observed signals, to resemble the noise-corrupted samples seen during training. An unconditional model then denoises the entire sequence, generating both the observed and missing values jointly. \textit{Self-Guidance} ~\cite{kollovieh2024predict} conditions on arbitrary patterns of observed values by leveraging the unconditional model's own estimates. Unlike classifier guidance, which relies on a separately trained classifier for likelihood estimation, self-guidance repurposes the denoiser itself for this task. It first computes a rough signal estimate  $\hat{\mathbf{x}}^{(:)}_0$ from the current outputs $\hat{\mathbf{x}}^{(:)}_t$, by inverting the forward noising process in eq.~\eqref{eq:ffwdnoising}. The samples are then iteratively corrected using the gradient of the conditional log-likelihood of the observed signals directly at inference.



\begin{figure*}[t]
    \centering
    \begin{subfigure}{0.3\textwidth}
    \includegraphics[width=\textwidth]{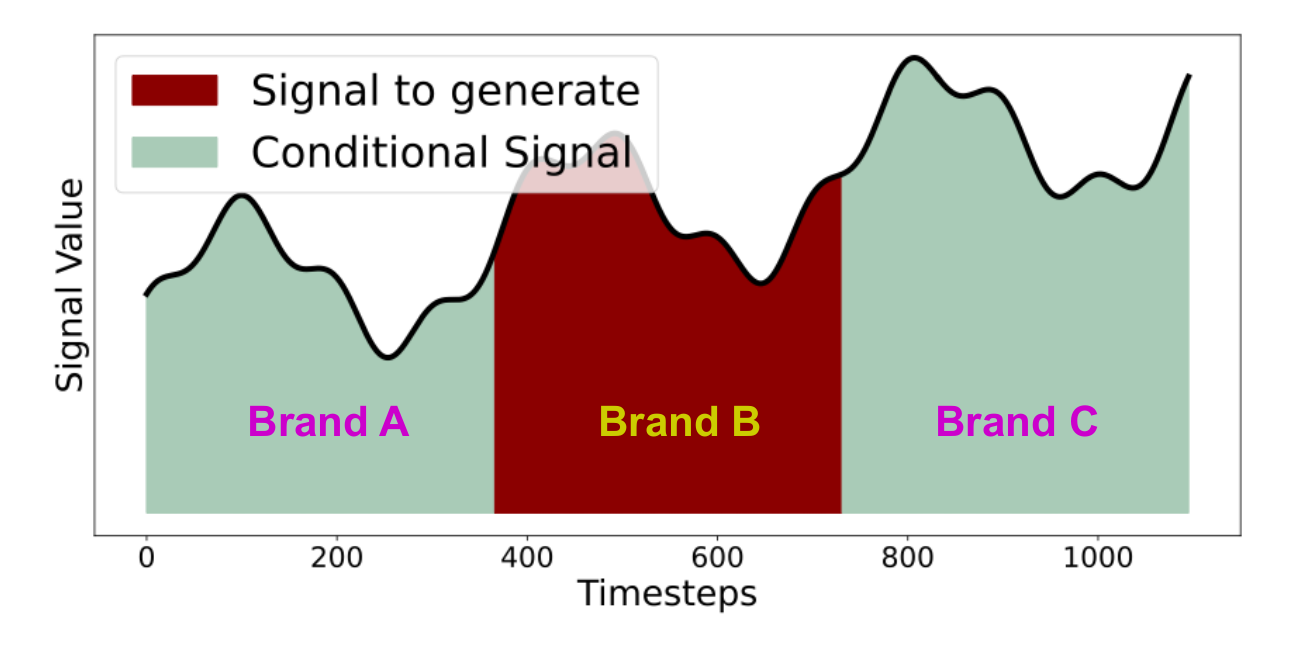} 
    \caption{Root: $(B, *, *)$}
    \label{fig:year}
    \end{subfigure}
    \begin{subfigure}{0.3\textwidth}
    \includegraphics[width=\textwidth]{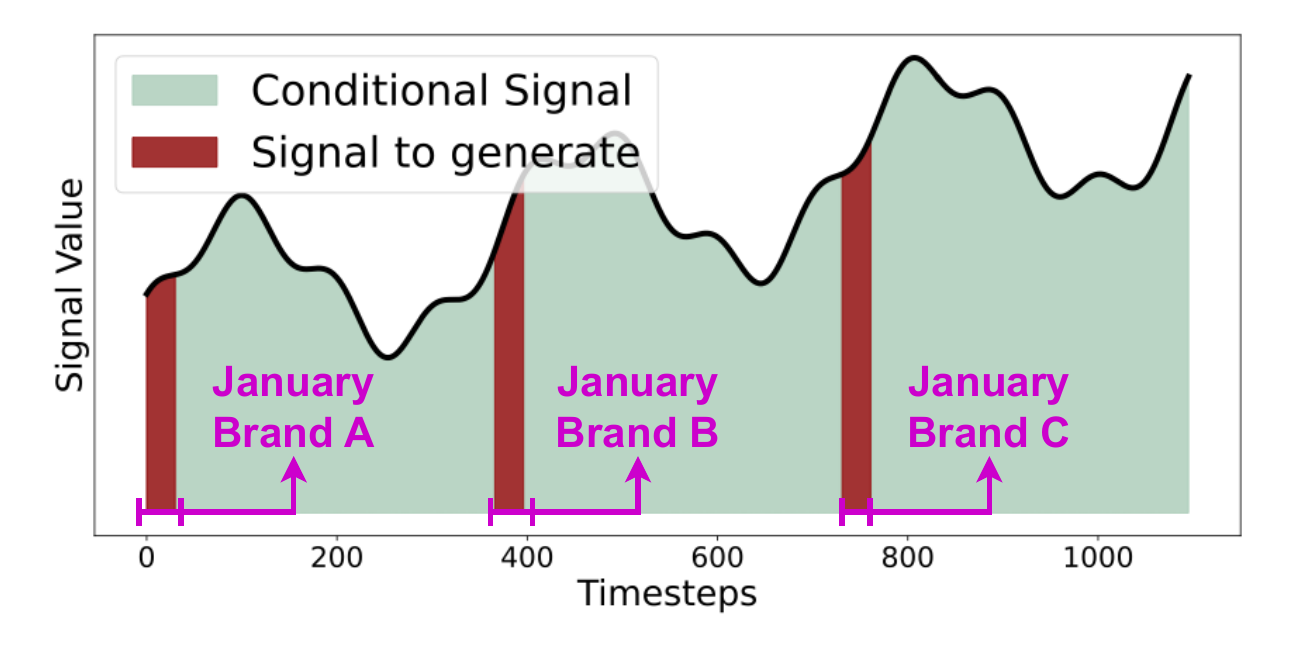}  
    \caption{Intermediate: $(*, {January}, *)$}
    \label{fig:month}
    \end{subfigure}
    \begin{subfigure}{0.3\textwidth}
    \includegraphics[width=\textwidth]{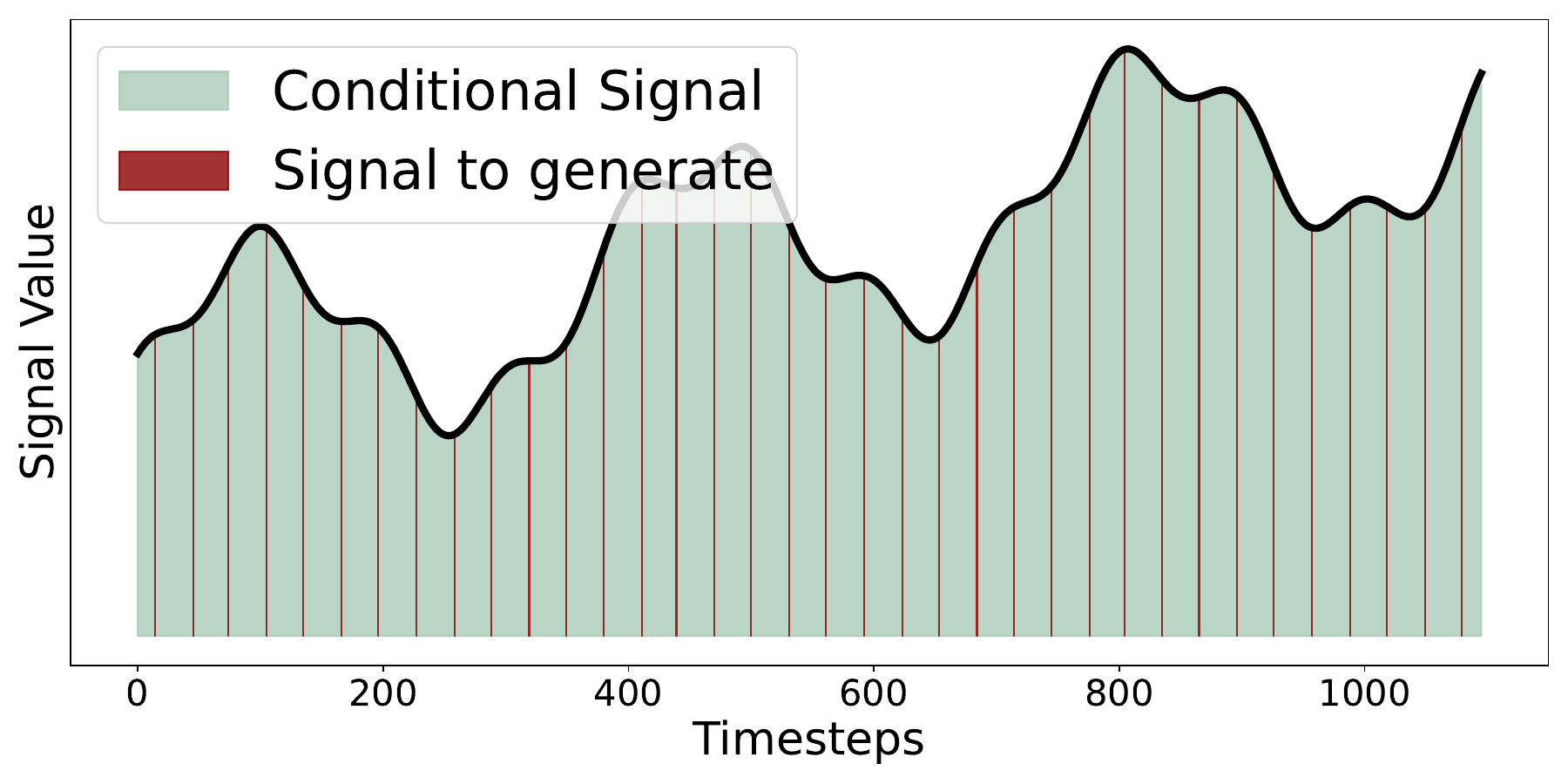}
    \caption{Bottom: $(*, *, 15)$}
    \label{fig:day}
    \end{subfigure}
    \caption{(a) Root (coarse-grained), (b) Intermediate and (c) Bottom level (fine-grained) conditions in \hyperref[exmp:2]{Example 1}.}
    \label{fig:constraintfigure}
\end{figure*}

\label{subsec:dynamic}




{\color{black}\subsection{Problem Definition}
\label{subsec:pd}
We are given a time series dataset. 
\(
\mathcal{D} = \{(\mathbf{a}^{(i)}, \mathbf{x}^{(i)})\}_{i=1}^N
\) with $N$ timesteps, 
where $\mathbf{a}^{(i)}\in \mathbb{R}^{1\times L}$ denotes a sample with $L$ metadata features and $\mathbf{x}^{(i)}\in \mathbb{R}^{1\times C}$ is a $C$-variate signal at the $i^{\text{th}}$ timestep.  
The aim is to generate time series signals given the metadata and signals observed at arbitrary timesteps. Training inputs include complete pairs of metadata and signals, i.e., training uses signals $\mathbf{x}^{(i)}$ with available metadata $\mathbf{a}^{(i)}$.
The model must generate time series signals conditioned on the metadata and the observed signals \textit{at inference time}. Conditional metadata may be specified as $\mathbf{a}^{(k)} = (\text{u}, \text{v}, *, *, *)$, indicating that signals should be generated (or possibly re-generated) at all timesteps $k$ where the first two metadata columns match $\text{u}$ and $\text{v}$ respectively. The asterisk ($*$) indicates unconstrained features. Therefore, the set 
$\{\mathbf{a}^{(k)}\}$ is assumed to contain all valid enumerations of the unspecified metadata features that satisfy the given condition. These enumerations are then inserted into 
$\mathcal{D}$ at their appropriate positions in the timeline. The model then needs to generate signals $\mathbf{x}^{(k)}$ conditioned on $\mathbf{a}^{(k)}$ and any observed signals $\mathbf{x}^{(i \neq k)}$ already present in $\mathcal{D}$. We further illustrate this problem setup using \hyperref[exmp:2]{Example 1} and \autoref{fig:constraintfigure}.

\begin{myframe}[innerleftmargin = 5pt]
\label{exmp:2}
\bemph{Example 1.} Consider a dataset containing one year’s worth of daily sales data for three brands, \textit{A}, \textit{B}, and \textit{C}. Each Sales entry (the signal) corresponds to a particular \emph{Brand}, on a given \emph{Month}, and \emph{Day}. The data is organised by concatenating the entries from all three brands into a table where each row is assigned a timestep index based on its position. 

A condition on the root-level metadata feature, such as \( (\text{\emph{Brand}}=B, *, *) \), requires generating sales for the entire year for Brand \textit{B}. As \autoref{fig:year} shows, the generated portions comprise a long \emph{contiguous block} of timesteps, resembling  \emph{coarse-grained} synthesis. In contrast, a condition on the bottom-level metadata feature, such as \((*, *, \text{\emph{Day}}=15) \), generates sales for the $15^{\text{th}}$ day of each month and brand, resembling \emph{fine-grained} tasks, as shown in \autoref{fig:day}. Similarly, tasks may also have intermediate granularity by conditioning mid-level features, like in \autoref{fig:month}
.
\end{myframe}

\begin{table}[tb]
\caption{Summary of key symbols.}
\label{tab:notation}
\centering
\scalebox{0.90}{
\begin{tabular}{@{}l @{\hspace{1cm}} l@{}}
\toprule
\textbf{Symbol}        & \textbf{Description} \\ 
\midrule
$\mathbf{x}^{(:)}_t, \mathbf{\hat{x}}^{(:)}_t$ & Ground-truth signal and model estimate at the $t$\textsuperscript{th} diffusion step \\
$(\alpha_t, \bar{\alpha}_t), \sigma_t$ & Diffusion parameters, sampling variance\\
$f_\theta$ & Denoiser neural network\\
$\mathcal{X, A, M}$ & Signal, metadata, and mask datasets\\
$\mathcal{D}$ & Full dataset comprising metadata and signals\\
$\mathcal{X}_w, \mathcal{A}_w, \mathcal{M}_w$ & Windowed signal, metadata, and mask datasets \\
$N \text{ or }M, L, C, T$ & Number of timesteps, metadata features, signal channels, and diffusion steps\\
$w,s,b,\eta$ & Window size, stride, {\color{black}mini-batch} size, guidance coefficient\\

$\mathbf{x}^{(i)}, \mathbf{m}^{(i)}, \mathbf{a}^{(i)}$ & Signal (ground-truth), mask, and metadata at the $i$\textsuperscript{th} timestep\\
$\mathbf{x}_w^{(j)}, \mathbf{m}_w^{(j)}, \mathbf{a}_w^{(j)}$ & $j$\textsuperscript{th} signal window (ground-truth), mask, and metadata window\\
$\mathbf{\hat{x}}_{w,t}^{(j)}$ & $j$\textsuperscript{th} signal window (model estimate) at diffusion step $t$\\
$t_\theta(w)$ & Denoising time (with $f_\theta$) for a  window of size $w$\\

\bottomrule
\end{tabular}}
\end{table}

\subsection{Research Gaps}

The inability to generalise to unseen conditional patterns limits training time methods. For instance, TimeWeaver \cite{narasimhan2024time} conditions only on the metadata and cannot leverage information from the observed signals. Similarly, SSSD~\cite{alcaraz2022diffusion} requires knowing the conditional mask of observations, which can lead to overfitting on the patterns seen during training. Classifier-free guidance requires both a conditional and unconditional model variant, which still requires training the conditional one with specific conditions in mind~\cite{ho2021classifier}. Even classifier guidance depends on pre-trained classifiers tailored to each condition type~\cite{liu2024controllable}, making it inflexible during inference since the classifiers require training with the right conditions in advance.

Among inference-time methods, RePaint separately adds noise to the observed and missing portions, causing misalignment between the two~\cite{lugmayr2022repaint,chung2022improving}. These inconsistencies are further amplified in time series, where maintaining coherence across time windows is also critical. Hence, RePaint's effectiveness in image inpainting \cite{lugmayr2022repaint} does not carry over to time series generation. Although ImDiffusion~\cite{imdiffusion} applies RePaint for time series, it focuses on anomaly detection rather than imputation or generation and also does not consider metadata. Gradient-based inference-time conditioning using self-guidance~\cite{kollovieh2024predict} improves upon RePaint by jointly updating known and unknown regions. However, it ignores metadata and conditions only on the observed signals within a given window, limiting its use for maintaining coherence across time windows.}

\section{\algo}
\label{subsec:wave}

This section first gives an overview of our approach and the preprocessing step for encoding categorical features (Sec.~\ref{ssec:prepro}). We then detail the core components of \algo (Sec.~\ref{subsec:tr}, Sec.~\ref{subsec:inf}): {\color{black}a \textit{dual-sourced} conditioning strategy that \textit{hybridises} training and inference approaches by combining a \textit{metadata-conditioned} model during training with the observed signals at inference. We also theoretically analyse the role of the inference-time \textit{conditional loss} (Sec.~\ref{subsec:theory}), to understand how it guides the sampling trajectory to produce coherent, observation-aligned outputs.}

\subsection{Overview}
Real-world time series tasks often involve conditioning on partial and irregular information, such as signal observations at arbitrary timesteps. A generative model must handle static metadata and arbitrarily-positioned signal observations to operate under these settings. As illustrated in \hyperref[exmp:2]{Example 1} and \autoref{fig:constraintfigure}, the observed (in \textcolor{darkseagreen}{green}) and missing (in \textcolor{darkred}{red}) signals can vary in position and size depending on the metadata conditions. Since this structure is not known a priori at inference, {\color{black}\algo hybridises training-time \textit{and} inference-time conditioning to enable dual-source conditioning for adapting to a wide range of conditional patterns.} 

We first train a \textit{metadata-conditioned} denoiser $f_{\theta}$ to predict noise using \textit{only} metadata, without relying on specific patterns of observed signals. This approach enables \algo to generate reasonable signal estimates even in the worst-case scenario, where we are given only the metadata and no observed signals as conditions. \algo then iteratively refines the generated samples during inference via a \textit{gradient-based} correction. At each denoising step, we adjust the generated sample using the gradient of a novel \textit{conditional loss} computed over observed signals. This approach is similar to \textit{gradient inversion attacks}~\cite{jeon2021gradient}, where an attacker iteratively optimises a dummy input using gradients to retrieve the original data while keeping the model frozen.

\subsection{Preprocessing Categorical Metadata}
\label{ssec:prepro}

Time series metadata often contains recurring categorical attributes, such as day of the week, month, and hour. 
To capture the periodic structure in these features, we use sine and cosine transformations inspired by positional encodings in Transformers~\cite{vaswani2017attention} and related work in time series generation~\cite{suh2024timeautodiff}. For a given feature, we map each category \( k \in \{0, 1, \dots, K-1\} \) to an angle \( \theta_k = \frac{2 \pi k}{K} \) on the unit circle. Each categorical value \( k \) is then encoded by the coordinates \( (\sin(\theta_k), \cos(\theta_k)) \), reflecting the feature’s cyclical nature. This representation is more compact than \textit{one-hot encoding} ~\cite{rodriguez2018beyond,kotelnikov2023tabddpm} and represents periodic behaviour more naturally.

\subsection{Training Metadata-conditioned Denoiser}
\label{subsec:tr}

\begin{figure}[tb]
    \centering   \includegraphics[width=\textwidth]{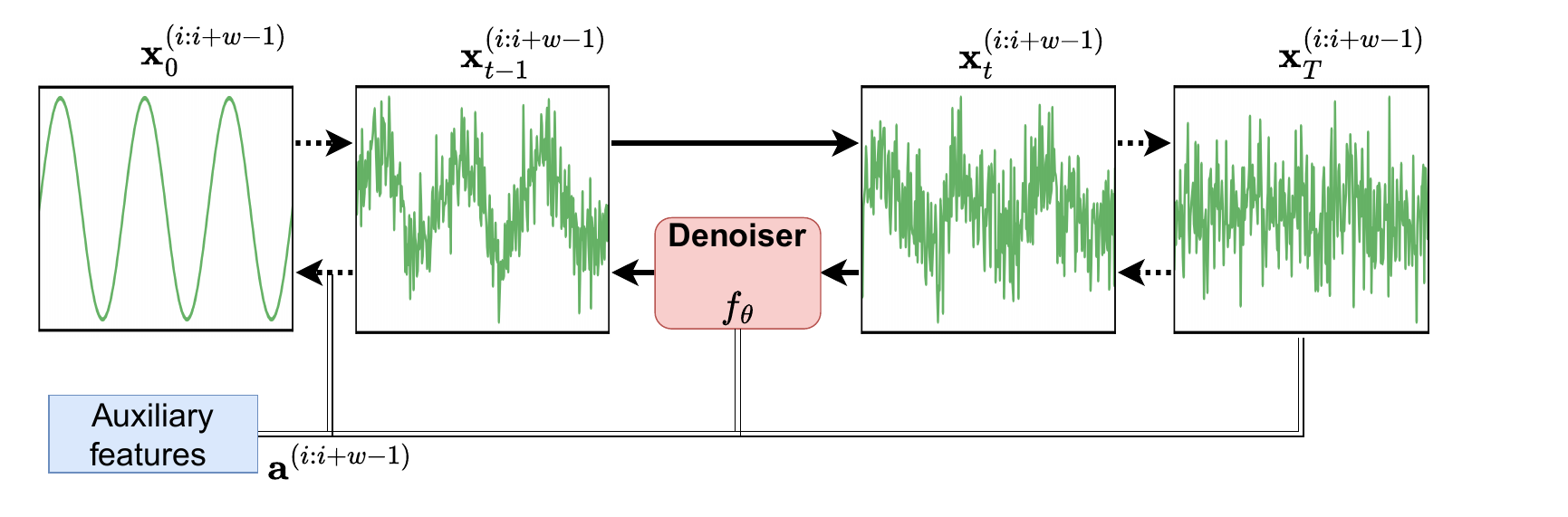}
    \caption{Metadata-conditioned denoiser training.}
    \label{fig:training}
\end{figure}

 We illustrate our training process in \autoref{fig:training} and show the steps in \hyperref[alg:training]{Algorithm 1}. It begins with dividing the entire dataset with $N$ timesteps into overlapping windows of size $w$ (line \ref{alg:trainingwindow}), using a stride of one. We then forward the noise up to a randomly sampled diffusion step $t$ and estimate the added noise, following the standard DDPM training process~\cite{ho2020denoising} (lines \ref{ln:fwdnoise}, \ref{ln:estimatenoise}). The noise is estimated by conditioning on the metadata for the window, $\mathbf{a}^{(i:i+w-1)}$, and uses the MSE loss to update $f_\theta$'s parameters (line \ref{ln:loss}). This way, the model learns the interactions between the metadata and the signals without relying on any observed signals. This separation sets the foundation to flexibly condition on the observations during inference (challenge 2). As we showed in \autoref{fig:day}, leveraging the observed values can further improve generation quality, particularly for fine-grained tasks where the observed signals from neighbouring timesteps can yield higher precision.

\begin{algorithm}
\caption{Training the Denoiser \( f_{\theta} \) Conditioned on Metadata}
\label{alg:training}
\begin{algorithmic}[1]

\State \textbf{Input:} Encoded Metadata \( \mathcal{A} = \{ \mathbf{a}^{(i)} \}_{i=1}^{N} \); Multivariate time series data \( \mathcal{X} = \{ \mathbf{x}^{(i)} \}_{i=1}^N \); Window size \( w \); Maximum number of denoising steps \( T \); Noise schedule \( \bar{\alpha} = \{ \bar{\alpha}_t \}_{t=1}^T \)

\For{each epoch}
    \For{\(i\) in range \(1\) to \(N-w+1\)}
        \State \textbf{Extract windows:} \( \mathbf{x}^{(i:i+w-1)} \) and \( \mathbf{a}^{(i:i+w-1)} \) of size \( w \).
        \label{alg:trainingwindow}
        \State \textbf{Sample} \( t \sim U(1,T) \) and \( \epsilon^{(i:i+w-1)} \sim \mathcal{N}(0, I) \)
        \State \textbf{Forward noise}:
        \label{ln:fwdnoise}
       \[
          \mathbf{x}_t^{(i:i+w-1)}= \sqrt{\bar{\alpha}_{t}} \mathbf{x}^{(i:i+w-1)} + \sqrt{1 - \bar{\alpha}_{t}} \epsilon^{(i:i+w-1)} 
        \]
        \vspace{-15pt}
        \State \textbf{Estimate} noise:
        \label{ln:estimatenoise}
        \[
        \hat{\epsilon}^{(i:i+w-1)} = f_{\theta}(\mathbf{a}^{(i:i+w-1)}, \mathbf{x}_t^{(i:i+w-1)}, t)
        \]
        \vspace{-15pt}
    \EndFor
    \State \textbf{Compute loss}:
    \label{ln:loss}
    \vspace{-5pt}
    \[
    \mathcal{L} = \mathbb{E}_{i \sim N-w+1}  \left\| \epsilon^{(i:i+w-1)} - \hat{\epsilon}^{(i:i+w-1)} \right\|^2 
    \]
    \vspace{-10pt}
    \State \textbf{Update} \( f_{\theta} \) parameters using \( \mathcal{L} \).
\EndFor
\end{algorithmic}
\end{algorithm}

\subsection{Signal-conditioned Inference}
\label{subsec:inf}
With our metadata-conditioned denoiser (Sec.~\ref{subsec:tr}), we additionally condition on the observed signals that are injected directly at inference, which we detail as follows:

\subsubsection{Pre-processing for data windows.}  We define the test set \( \mathcal{X} = \{ \mathbf{x}^{(i)} \}_{i=1}^M \), where \( \mathbf{x}^{(i)} \) indicates the observed signals, and the unknown values are left empty, representing the portions to be generated. The complete set of metadata, \( \mathcal{A} = \{ \mathbf{a}^{(i)} \}_{i=1}^M \) is available as part of the setup, consistent with the problem definition (Sec.~\ref{subsec:pd}). A corresponding mask \( \mathcal{M} = \{ \mathbf{m}^{(i)} \}_{i=1}^M \) is defined, where \( \mathbf{m}^{(i)} = 1 \) indicates that \( \mathbf{x}^{(i)} \) requires generation and \( \mathbf{m}^{(i)} = 0 \) indicates that the corresponding signals are observed. {\color{black}We first divide the test set into overlapping time windows of size \( w \), spaced by a stride \( s \), yielding \( \mathcal{X}_w = \{ \mathbf{x}_w^{(j)} \}_{j=1}^{\left\lceil \frac{M - w}{s} \right\rceil+1} \), \( \mathcal{M}_w = \{ \mathbf{m}_w^{(j)} \}_{j=1}^{\left\lceil \frac{M - w}{s} \right\rceil+1} \), and \( \mathcal{A}_w = \{ \mathbf{a}_w^{(j)} \}_{j=1}^{\left\lceil \frac{M - w}{s} \right\rceil+1} \)}. Following this, we proceed with the generation process detailed in  \hyperref[alg:synth]{Algorithm 2}, which iterates through steps of $t=T,T-1 \dots 1$. At each step, the metadata-conditioned denoiser first produces an intermediate sample, which is then adjusted using conditional loss gradients.  

\begin{algorithm}
\caption{Parallel Denoising with Inference Conditioning}
\label{alg:synth}
\begin{algorithmic}[1]
\State \textbf{Input:} Windowed metadata ($\mathcal{A}_w$), masks ($\mathcal{M}_w$), and signals ($\mathcal{X}_w$); Timesteps $M$, window size \( w \); Stride \( s \); mini-batch size \(b\); strength $\eta$; Diffusion parameters \( \{ \alpha_t, \bar{\alpha}_t\} \); denoiser \( f_\theta \).

\State \textbf{Initialize outputs:}
\vspace{-5pt}
\[
\mathcal{\hat{X}}_w = \{ \mathbf{\hat{x}}_{w,T}^{(j)}\sim \mathcal{N}(0,I) \}_{j=1}^{\left\lceil {(M - w)}/{s} \right\rceil+1} 
\] \label{ln:init}
\vspace{-13pt}
\State \textbf{Divide $\mathcal{X}_w, \mathcal{A}_w, \mathcal{M}_w$ into $(M-w)/(b\times s)$ mini-batches}\label{ln:mini}

\For{each mini-batch}
  \For{(\(\mathbf{x}_w^{(j)}, \mathbf{a}_w^{(j)}, \mathbf{m}_w^{(j)}\)) in mini-batch in \textbf{parallel}:}
    \For{step \( t = T, T-1, \dots, 1 \)}

      \State \textbf{Dirty estimate }\(\mathbf{\hat{x}}_{w,0}^{(j)} \) \label{ln:ffden} using eq.~\eqref{eq:dirtyestimate} 
      \State \textbf{Unconditional denoising:} 
      \(\mathbf{\hat{x'}}_{w,t-1}^{(j)}\)  using eq.~\eqref{eq:initguess}\label{ln:uncond}
      \State \textbf{Compute conditional loss} \(\mathcal{L}_{\text{cond}}^{(j)}\) using eq.~\eqref{eq:finalcondloss} \label{ln:condlossline}
      \State \textbf{Gradient correction} of \(\mathbf{\hat{x'}}_{w,t-1}^{(j)}\) using eq.~\eqref{eq:onestepconditden}
      \label{ln:gradcorrection}
        
      \EndFor
      \State \textbf{Re-introduce observations:}
        \[
        \mathbf{\hat{x}}_{w,0}^{(j)} = (1-\mathbf{m}_w^{(j)})\cdot \mathbf{x}_w^{(j)}
         + \mathbf{m}_w^{(j)}\cdot \mathbf{\hat{x}}_{w,0}^{(j)}\] \label{ln:reint}
  \EndFor
 
\EndFor

\State \textbf{Merge windows:}
\vspace{-5pt}
\[
\mathcal{\hat{X}} = \mathbf{\hat{x}}_{w,0}^{(1)} \cup 
\left(
\bigcup_{j\geq 2} \mathbf{\hat{x}}_{w,0}^{(j)(w-s+1:w)}
\right)
\]
\vspace{-10pt} \label{ln:merge}
\State \Return {\(\mathcal{\hat{X}}\)}

\end{algorithmic}
\end{algorithm}
\subsubsection{Metadata-Conditioned Generation.}
We first initialise the signal to be generated ($\mathcal{\hat{X}}_w$) in line \ref{ln:init}. Then we divide $\mathcal{X}_w$, $\mathcal{M}_w$, and $\mathcal{A}_w$ into {\color{black}mini-batches} of size $b$, each containing samples of $w$ timesteps (line \ref{ln:mini}). To condition on the observed signals \textit{and} metadata, we first get a dirty estimate of the fully-denoised signal ($\mathbf{\hat{x}}_{w,0}^{(j)}$) using our metadata-conditioned denoiser (line~\ref{ln:ffden}). We get this estimate by reversing the fast-forward noising step in eq.~\eqref{eq:ffwdnoising} as follows:
\begin{equation}
\label{eq:dirtyestimate}
    \mathbf{\hat{x}}_{w,0}^{(j)} = 
    \frac{\mathbf{\hat{x}}_{w,t}^{(j)} - \sqrt{1-\bar{\alpha}_t} \cdot f_{\theta} \left( \mathbf{a}_w^{(j)},    \mathbf{\hat{x}}_{w,t}^{(j)},t \right)}{\sqrt{\bar{\alpha}_t}} 
\end{equation}

Here, $\mathbf{\hat{x}}_{w,t}^{(j)}$ is the model's estimate at the $t\textsuperscript{th}$ diffusion step.

\subsubsection{Sample adjustment with gradient inversion.}
The denoiser $f_{\theta}$ estimates samples by only conditioning on the metadata. To further factor in the observed signal values, we correct the samples using the gradients from the \textit{conditional loss} (line~\ref{ln:condlossline}), defined as follows:

\begin{equation}
\label{eq:finalcondloss}
\mathcal{L}_{\text{cond}}^{(j)} = \mathcal{L}_{\text{self}}^{(j)} + \mathcal{L}_{\text{stitch}}^{(j)}
\end{equation}
This loss consists of two components: the \textit{self-guidance} loss ($\mathcal{L}_{\text{self}}^{(j)}$), which enforces consistency with the observed signals \textit{within} each time window, and the \textit{stitch} loss ($\mathcal{L}_{\text{stitch}}^{(j)}$), which promotes coherence \textit{across} overlapping windows.

 \emph{\underline{Self-Guidance}:} The \emph{self-guidance loss}~\cite{kollovieh2024predict} ensures that the generated outputs match the observed signals within a given time window. It is defined as follows:

\begin{equation}
\label{eq:selfguidance}
\mathcal{L}_{\text{self}}^{(j)} = \|(1 - \mathbf{m}_w^{(j)}) \odot (\mathbf{\hat{x}}_{w,0}^{(j)} - \mathbf{x}_{w}^{(j)})\|^2
\end{equation}

Here, the term $1 - \mathbf{m}_w^{(j)}$ is used to compute the loss only on the observed timesteps, enforcing local consistency within a window.

\emph{\underline{Stitching}:} Since self-guidance treats each window independently, it may produce inconsistencies across window boundaries by ignoring overlaps. These overlaps are crucial for maintaining temporal continuity, and failing to align them can result in incoherent transitions. 
\textit{Stitching} overcomes this limitation 
by enforcing consistency in overlapping regions, restricting the solution space to a more realistic subset than those produced by self-guidance alone. For a given window \( \mathbf{\hat{x}}_{w,t}^{(j)} \), we first denoise one step \( \mathbf{\hat{x'}}_{w,t-1}^{(j)} \) \textit{unconditionally} using eq.~\eqref{eq:denoiseone} (line~\ref{ln:uncond}) :
\begin{equation}
    \label{eq:initguess}
    \mathbf{\hat{x'}}_{w,t-1}^{(j)} = \frac{1}{\sqrt{\alpha_t}} \left( \mathbf{\hat{x}}_{w,t}^{(j)} - \frac{1 - \alpha_t}{\sqrt{1 - \bar{\alpha}_t}} \cdot f_{\theta} \left( \mathbf{a}_w^{(j)}, \mathbf{\hat{x}}_{w,t}^{(j)},t \right) \right)  
    + \sigma_t z
\end{equation}

We then \textit{``stitch''} these uncorrected estimates across window boundaries to ensure coherence.   Specifically, for a window of size \( w \) and stride \( s \), the overlap consists of the first \( w - s \) timesteps of \( \mathbf{\hat{x'}}_{w,t-1}^{(j)} \) and the last \( w - s \) timesteps of \( \mathbf{\hat{x'}}_{w,t-1}^{(j-1)} \). The stitch loss then penalises discrepancies in this overlap as follows\footnote{For the first window, we define \( \mathcal{L}_{\text{stitch}}^{(1)} = 0 \), as there are no preceding windows.}:

\begin{equation}
\label{eq:stitching}
\mathcal{L}_{\text{stitch}}^{(j)} = \|\mathbf{\hat{x'}}_{w,t-1}^{(j)(1:w-s)} - \mathbf{\hat{x'}}_{w,t-1}^{(j-1)(1+s:w)}\|^2
\end{equation}

\emph{\underline{Adjustment}:} To dynamically condition on the observed signals, we correct the unconditional outputs from eq.~\eqref{eq:initguess} using the conditional loss gradients scaled by guidance coefficient \(\eta\) (line~\ref{ln:gradcorrection}): 

\begin{equation}
\label{eq:onestepconditden}
    \mathbf{\hat{x}}_{w,t-1}^{(j)} = \mathbf{\hat{x'}}_{w,t-1}^{(j)}  
     - \eta \nabla_{\mathbf{\hat{x}}_{w,t}^{(j)}} \mathcal{L}_{\text{cond}}^{(j)},
\end{equation}

 where $\mathcal{L}_{\text{cond}}$ is as defined in eq.~\eqref{eq:finalcondloss}, $z\sim \mathcal{N}(0,I)$, and $\sigma_t = (1-\alpha_t)\times\frac{1-\bar{\alpha}_{t-1}}{1-\bar{\alpha}_t}$. At the end of the denoising process, we reintroduce the observations (line \ref{ln:reint}) and merge the overlapping windows (line \ref{ln:merge}), to yield the final sequence. \autoref{fig:stitch} illustrates the conditional denoising process using the combination of self-guidance and stitching. 

 {\color{black}To summarise, eq.~\eqref{eq:initguess} unconditionally denoises the sample without enforcing any constraints. If left unmodified, the reverse diffusion trajectory would follow this path, ignoring both observations and inter-window coherence. Equation~\eqref{eq:onestepconditden} adjusts the trajectory using the gradient of the conditional loss, which consists of two terms. The self-guidance loss eq.~\eqref{eq:selfguidance} enforces alignment with the observed signal values \textit{within} each window, while the stitching eq.~\eqref{eq:stitching} enforces consistency \textit{across} overlapping windows. Together, these components nudge the model at every denoising step to yield samples that adhere to observations and remain globally coherent.

{\color{black}
\subsection{Theoretical Analysis}
\label{subsec:theory}
Here, we formalise how the inference-time conditional loss, comprising self-guidance and stitching, constrains generation towards a tighter solution space that promotes global temporal coherence and alignment with the given signal observations.

\textbf{Definitions.} Let \( {\mathcal{X}_{\text{cond}}^{\delta}} \) be the set of samples for which the conditional loss is at most $\delta$ $\ge 0$, i.e., they respect both the overlap consistencies and the observed signals within windows up to a tolerance $\delta$. As a special case, \( {\mathcal{X}_{\text{cond}}^{0}} \) represents the set of samples with zero-conditional loss. Similarly, let \( {\mathcal{X}_{\text{stitch}}^{\delta}}\) and \( {\mathcal{X}_{\text{self}}^{\delta}}\) be the set of samples with at most $\delta$ stitch and self-guidance losses respectively. Finally, we define \( \mathcal{X}_{\text{real}} \) as the set of real samples, which must obey the following conditions: (a) exactly match the known values, i.e., zero self-guidance loss; and (b) be temporally coherent, i.e., any overlapping windows yield identical values on their shared timesteps (zero stitch loss). A violation of condition (b) would result in conflicting predictions at the same timestep upon merging, which is not possible for a real time series.

\begin{proposition}
With sets \( \mathcal{X}_{\text{real}} \), \( \mathcal{X}_{\text{self}}^\delta \), \( \mathcal{X}_{\text{stitch}}^\delta \), and \( \mathcal{X}_{\text{cond}}^\delta\)  defined as above, we have the following chain of inclusions:
\[
\mathcal{X}_{\text{real}} \subseteq \mathcal{X}_{\text{cond}}^0 \subseteq \mathcal{X}_{\text{cond}}^\delta \subseteq \mathcal{X}_{\text{self}}^\delta.
\]
\end{proposition}

\begin{proof}
By definition, the conditional loss is the sum of the self-guidance loss and the stitch loss, with all losses being non-negative (see eq.~\eqref{eq:finalcondloss}). Therefore, for any \( x \in \mathcal{X}_{\text{cond}}^\delta \), we have:
\[
\mathcal{L}_{\text{cond}}(x) \leq \delta \quad \Rightarrow \quad \mathcal{L}_{\text{self}}(x) \leq \delta \quad \text{and} \quad \mathcal{L}_{\text{stitch}}(x) \leq \delta.
\]
This implies that for all $ x \in {\mathcal{X}_{\text{cond}}^\delta}$:
\[
 x \in \mathcal{X}_{\text{self}}^\delta \cap \mathcal{X}_{\text{stitch}}^\delta \Rightarrow \mathcal{X}_{\text{cond}}^\delta \subseteq \mathcal{X}_{\text{self}}^\delta.\tag{I}
\]
Next, since the zero conditional loss is a special case of loss less than or equal to \(\delta\), we have:
\[
\mathcal{X}_{\text{cond}}^{0} \subseteq \mathcal{X}_{\text{cond}}^{\delta}. \tag{II}
\]

Finally, according to the earlier definition, any \( x \in \mathcal{X}_{\text{real}} \) must have zero self-guidance loss and zero stitch loss. Hence, for all $ x \in \mathcal{X}_{\text{real}}$, we have:
\[
x \in \mathcal{X}_{\text{self}}^0 \cap \mathcal{X}_{\text{stitch}}^0 = \mathcal{X}_{\text{cond}}^0 \Rightarrow \mathcal{X}_{\text{real}} \subseteq \mathcal{X}_{\text{cond}}^0.\tag{III}
\]
Combining (I), (II), and (III), we then obtain the chain of inclusions:
\[
\mathcal{X}_{\text{real}} \subseteq \mathcal{X}_{\text{cond}}^0 \subseteq \mathcal{X}_{\text{cond}}^\delta \subseteq \mathcal{X}_{\text{self}}^\delta.
\]
\end{proof}

\noindent
 Proposition 3.1 shows that the conditional loss restricts generation to a subset closer to the real data space. While self-guidance ensures agreement with the known values within each window, it does not constrain the unknown regions. As a result, overlapping windows may assign conflicting values to the same timestep, yielding inconsistent samples after merging. Stitching penalises inconsistencies in the overlaps. Hence, the sample set with zero conditional loss is a subset of the self-guided sample space and contains all realistic samples. Since the model trains on realistic and temporally coherent samples, the inference-time conditional loss does not push it away from the training distribution. Rather, it keeps samples within the real space by bringing the model’s outputs closer to the characteristics it was initially optimised to reproduce.

While zero conditional loss is the limit, the proposition implies that even minimising it to a small value $\delta > 0$ brings the samples closer to realistic behaviour, even without exact convergence. Crucially, zero conditional loss is a necessary but not a sufficient condition for realism. Hence, there may exist samples that perfectly match the observations with coherent overlaps but still diverge from the true sequence. This observation reflects an inherent ambiguity in conditional generation. If multiple completions can satisfy the given conditions, minimising the conditional loss improves plausibility, but cannot guarantee a match to the ground truth.
}



\section{Pipelined Parallel Design of \algo}
\label{sec:para}
\begin{figure*}[tb]
    \centering
      \includegraphics[width=\textwidth]{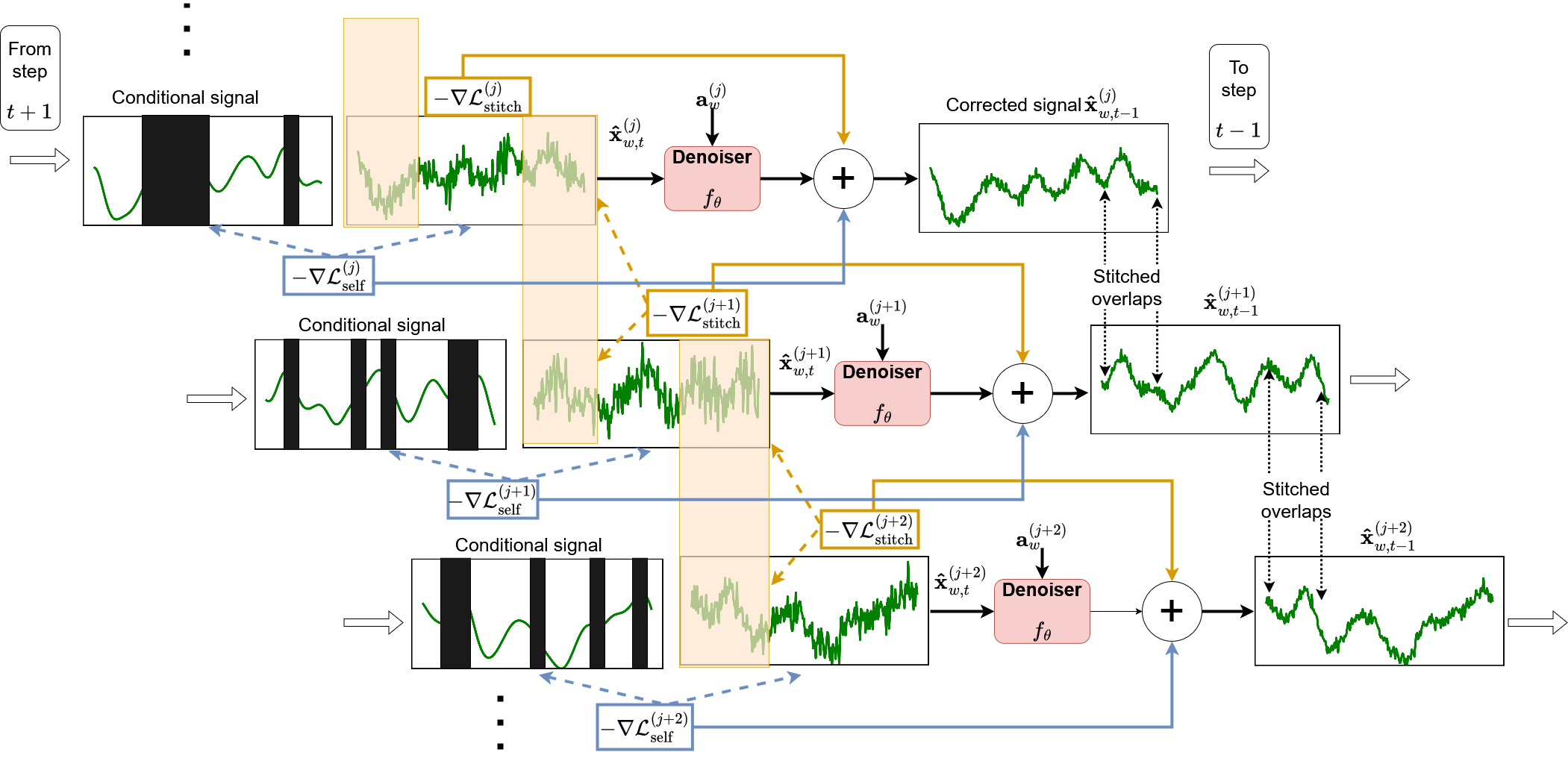}
    \caption{Parallel Inference Conditioning with Stitching and Self-Guidance Loss. The denoising process flows from left to right. At each step $t$, the model receives denoised windows from step $t+1$. It then refines these using two types of gradient corrections: the self-guidance loss (in \textcolor{colblue}{blue}), computed using observed signals, and the stitching loss (in \textcolor{colora}{orange}) on the overlaps. These corrections update the metadata-conditioned estimate before passing it to the next denoising step $t-1$.}
    \label{fig:stitch}
     \vspace{-0.4cm}
\end{figure*}

{
The conditioning mechanism described in Sec.~\ref{subsec:inf} could be applied autoregressively by processing each window sequentially. However, most steps in \hyperref[alg:synth]{Algorithm 2} can be performed independently across windows. Even the conditional loss computation (\hyperref[alg:synth]{Algorithm 2}, line \ref{ln:condlossline}) can be made parallel-compatible by a parallel \textit{shift} (or \textit{roll}) operation to align overlapping regions, to compute the pairwise loss between consecutive windows concurrently. This process resembles a \emph{pipeline}, as shown in \autoref{fig:stitch}. At each denoising step, non-overlapping parts adjust to maintain coherence with their neighbours. Although these adjustments are only enforced with the neighbouring window, coherence gradually propagates over the whole pipeline over successive steps.

The stride \( s \) is crucial in balancing efficiency and coherence. When \( s \) is large enough to eliminate overlap, the process becomes a \emph{divide-and-conquer} strategy~\cite{li2024survey,yin2023nuwa}, that synthesises windows independently. Conversely, smaller strides enforce stronger dependencies between windows due to larger overlaps. In the extreme case of \( s = 1 \), maximum coherence is achieved as nearly all timesteps are shared between consecutive windows. However, the optimal stride depends on the time series itself. Smaller strides ensure better alignment for signals with strong temporal dependencies, while larger strides may suffice for weaker ones. This flexibility allows our method to adapt to varying degrees of temporal dependence for efficient and coherent signal generation.


\textcolor{black}{\textbf{Time Complexity and Speedup:}
\label{subsubsec:speed}
\algo substantially improves the generation speed compared to autoregression due to parallelism. Using the notation in \hyperref[alg:synth]{Algorithm 2}, each dataset has around \(\frac{M - w}{s} \) time windows, where \( M \) is the total length, \( w \) is the window size, and \( s \) is the stride. With sequential execution, the total cost is  \( \mathcal{O}(T \cdot t_\theta(w) \cdot (M-w)/s) \), where $t_{\theta}(w)$ is the time taken to denoise a window by one step. In contrast, the parallel approach processes \( b \) windows in parallel per {\color{black}mini-batch}, which significantly reduces the time complexity to \( \mathcal{O}(T \cdot t_\theta(w) \cdot \frac{M-w}{b \cdot s}) \). This analysis suggests an \emph{ideal speedup} factor of approximately $b$ over the autoregressive approach, representing an upper bound on performance under optimal hardware utilisation.}

\section{Evaluation}
\label{sec:eva}
{
We conduct a comprehensive set of experiments to evaluate how well \algo addresses the following research questions: \\
\textbf{RQ1:} How effective is \algo's dual-source and hybrid conditioning strategy in integrating metadata and observed signals compared to SOTA methods?\\
\textbf{RQ2:} How does the speed and quality of parallel synthesis compare to autoregressive generation?\\
\textbf{RQ3:} Does stitching improve performance compared to using only self-guidance or RePaint~\cite{lugmayr2022repaint,imdiffusion}?\\
\textbf{RQ4:} What is the impact of alternative conditional loss formulations on the generation quality?
\textbf{RQ5:} Does \algo generalise to random imputation tasks with varying missingness ratios?}

\textbf{System Configuration.}
We use a 12-core AMD Ryzen 9 5900 processor with an NVIDIA RTX 3090 GPU. For software, we use CUDA 12.1, Ubuntu 20.04.6, and Pytorch 2.2.1 with Python 3.12.  

\begin{table}[tb]
\centering
\caption{Datasets and Experiment Setup. For BeijingAirQuality (BQ), AustraliaTourism (AT), MetroTraffic (MT), PanamaEnergy (PE), and RossmanSales (RS). Length refers to the sequence length of the test set, equalling the number of samples for the \bemph{R}-level condition.}
\renewcommand{\arraystretch}{0.85} 
\large 
\scalebox{0.85}{ 
\begin{tabular}{lllll}
\hline
\textbf{Dataset} & \textbf{Metadata Features} & \textbf{Channels} & \textbf{Length} & \textbf{Condition}           \\ \hline
BQ               & Year, Station,     & 11                & 8496            & \bemph{R}: $(2017, *, *, *, *)$        \\
                  & Month, Day,        &                   &                 & \bemph{I}: $(2017, *, 2, *, *)$        \\
                  & Hour               &                   &                 & \bemph{B}: $(2017, *, *, *, 11)$       \\ \hline
MT                & Year, Month,       & 5                 & 7949            & \bemph{R}: $(2018, *, *, *)$           \\
                  & Day, Hour          &                   &                 & \bemph{I}: $(2018, *, 15, *)$            \\
                  &                    &                   &                 & \bemph{B}: $(2018, *, *, 6)$             \\ \hline
PE                & Year, Month,       & 4                 & 12819           & \bemph{R}: $(2020, *, *, *, *)$          \\
                  & Day, Hour,         &                   &                 & \bemph{I}: $(2020, *, 5, *, *)$          \\
                  & City               &                   &                 & \bemph{B}: $(2020, *, *, *, San)$        \\ \hline
RS                & Year, Month,       & 2                 & 1757            & \bemph{R}: $(2015, *, *, *)$             \\
                  & Day, Store         &                   &                 & \bemph{I}: $(2015, 3, *, *)$             \\
                  &                    &                   &                 & \bemph{B}: $(2015, *, *, 9)$             \\ \hline
AT                & Year, Month,       & 1                 & 1232            & \bemph{R}: $(2016, *, *, *, *)$          \\
                  & State, Region,     &                   &                 & \bemph{I}: $(2016, *, Queensland, *, *)$ \\
                  & Purpose            &                   &                 & \bemph{B}: $(2016, *, *, *, Holiday)$    \\ \hline
\end{tabular}}
\label{tab:datasets}
\vspace{-0.3cm}
\end{table}

\subsection{Datasets and Evaluation Framework}
\label{subsec:evalframework}
\textbf{Datasets.} We use public multivariate and univariate datasets from diverse domains: BeijingAirQuality (BQ)~\cite{beijing_multi-site_air_quality_501}, MetroTraffic Volume (MT)~\cite{metro_interstate_traffic_volume_492}, PanamaEnergy (PE)~\cite{panama_energy}, RossmanSales (RS)~\cite{rossmann-store-sales}, and AustraliaTourism (AT)~\cite{australia-tourism}. Each dataset includes metadata, grouped in the order specified in \autoref{tab:datasets}, reflecting how the data is organised. The time series channels for each dataset are as follows. For BQ, we use all gas sensor readings, temperature, pressure, dewpoint, rainfall, and wind speed from the first six stations, sorted alphabetically; for MT, we use temperature, rainfall, snowfall, traffic volume, and cloud cover levels; for PE, we use energy data such as T2M, QV2M, TQL, and W2M; for RS, we use sales and customer totals from stores 1-10; and for AT we use the number of trips.

\textbf{Evaluation Tasks.} Designing evaluation scenarios for our problem is not trivial. Random or standard 80-20 splits would disrupt temporal dependencies. To address this, we place all samples with a shared root-level value (e.g., a specific year) in the test set, ensuring unseen metadata combinations at inference. This approach supports {coarse-grained} imputation tasks, as well as {fine-grained} or {mid-level} tasks, by specifying lower-level metadata features within the test set.

To evaluate the model's performance across different scenarios, we use a structured approach that conditions metadata at three levels: \textit{Root} (\emph{R}), \textit{Intermediate} (\emph{I}), and \textit{Bottom} (\emph{B}), each corresponding to a different level in the grouping hierarchy and reflecting varying degrees of missing signal information and task granularity (see \autoref{tab:datasets}). Root Level \emph{(R)} conditions fix the root-level metadata feature (e.g. \textit{Year} in a \textit{Year}->\textit{Month}->\textit{Day} hierarchy), creating tasks missing contiguous signal blocks (e.g., \autoref{fig:year}). Intermediate Level \emph{(I)} tasks restrict mid-level metadata features, giving missing blocks of intermediate size (e.g., \autoref{fig:month}). Finally, Bottom Level \emph{(B)} tasks restrict the last metadata feature, yielding fine-grained imputation scenarios (e.g., \autoref{fig:day}). As shown in \autoref{tab:datasets}, both \textit{I} and \textit{B} additionally condition on the root-level feature to ensure that samples remain within the test set. While these configurations represent only a subset of possible scenarios, they cover a broad range of missingness patterns, allowing us to evaluate the model under various conditions. For example, this setup can also capture random missingness by combining multiple \emph{(B)} conditions.

\textbf{Metrics.} 
{\color{black}Imputation and generation are fundamentally different tasks and require different evaluation metrics. We treat forecasting as a special case of imputation, where all missing values appear only at the tail end of the time series, as discussed in Sec. 1 and explored in prior work \cite{alcaraz2022diffusion,kollovieh2024predict}. Imputation assumes the presence of \textit{a} ground-truth signal and focuses on accurately recovering missing values. Hence, the Mean Squared Error (MSE) is apt here, as it measures a point-wise match between predictions and ground-truths. In contrast, generation allows for multiple plausible outputs and aims for high-level resemblance with the real signals. Hence, metrics that assess similarity at a more abstract or distributional level are more appropriate here. These include statistics like feature-wise correlations or differences in autocorrelation, which capture structural or periodic resemblance between real and generated signals, rather than strict one-to-one matches. Therefore, we use the \emph{autocorrelation difference} (ACD)~\cite{wiese2020quant,xu2020cot} and \emph{cross-feature correlations} ($x$-Corr)~\cite{jarrett2021time}, in line with previous studies~\cite{ang2023tsgbench,ni2021sig,liao2024sig}. These capture higher-level similarities between the generated and ground-truth signals, such as similar {periodic patterns} (ACD) or {inter-feature dependencies ($x$-Corr)}.} Each metric's score is averaged over {five} trials to ensure robustness. 

We compute the ACD between the real and synthetic signal across multiple lags per channel. Let $ACF_{real}^{(c)}(\tau)$ and $ACF_{synth}^{(c)}(\tau)$ represent the autocorrelation values~\cite{park2018fundamentals} for lag $\tau$ in channel $c$ for the real and synthetic data respectively. Given the maximum number of lags $G$ and signal channels $C$, we can compute the difference between the real and generated autocorrelation values~\cite{wiese2020quant}:
\begin{equation}
    ACD = \frac{1}{C \cdot G} \sum_{c=1}^{C} \sum_{\tau=1}^{G} \left| ACF_{real}^{(c)}(\tau) - {ACF}_{{synth}}^{(c)}(\tau) \right|
\end{equation}
The ACD measures how similarly the generated and original signals evolve by comparing their autocorrelations across multiple time lags. For the cross-feature correlations, we calculate the pairwise Pearson correlation coefficients between all pairs of channels for both the real and synthetic time series. We then compute the average absolute difference in these correlations between the real and synthetic data, as done in prior works~\cite{jarrett2021time,xu2020cot,ni2021sig}.

\subsection{Model Implementation and Baselines}
We extend the implementations of several baselines from \autoref{tab:relwork} for conditional generation. For all diffusion models, including ours, we use SSSD~\cite{alcaraz2022diffusion} as the denoiser backbone for consistency, which utilises structured state space layers (S4 layers~\cite{smith2022simplified}) and has outperformed transformer-based methods such as CSDI~\cite{tashiro2021csdi}. {\color{black}The architecture consists of 4 hidden layers, four residual layers, and uses residual, skip, and hidden dimensions of size 64. The model employs embeddings for diffusion steps, with dimensions of 32 (input), 64 (middle), and 64 (output). S4 layers are configured with a maximum sequence length of 100, a state dimension of 64, and are used bidirectionally with layer normalisation and no dropout.}

{\color{black}
To ensure fairness, we re-implement all diffusion baselines: SSSD \cite{alcaraz2022diffusion}, TimeWeaver \cite{narasimhan2024time}, TSDiff \cite{kollovieh2024predict}, and TimeAutoDiff \cite{suh2024timeautodiff}, using the same backbone architecture described above.
SSSD is trained to impute randomly masked signal rows conditioned on the observed rows and metadata. We uniformly sample a number of timesteps between 0 and the total sequence length to randomly mask out during training. The model then trains to reconstruct the masked signals using the observed rows and the metadata. TimeWeaver, by design, conditions solely on metadata. To match this, we train it by masking out all signal rows, so that it learns to generate signals from just the metadata. We extend TSDiff to condition on metadata using the same metadata-conditioned denoiser as TimeWeaver. The observed signals are then conditioned on directly at inference time using self-guidance (see Sec.~\ref{subsec:conditionalddpm}). We tried different scales for tuning TSDiff's guidance strength: 0.0, 0.5, 1.0, and 2.0, and report results for 0.5, since it had the best overall performance. For fairness, we adapt TimeAutoDiff from its original latent diffusion architecture to operate in real space using our shared backbone. It uses classifier-free guidance that conditions on labels (see Sec.~\ref{subsec:conditionalddpm}), interpolating using two model variants. We adapt it to conditions on both the metadata and signal observations. One branch uses only metadata while the other conditions on both metadata and signal observations, like the SSSD baseline. We train each variant separately, and at inference, we interpolate between them using a guidance strength of 3.0, following prior work~\cite{ho2021classifier, suh2024timeautodiff}}.

We also compare against the GAN-based method TimeGAN \cite{yoon2019time}, which consists of five networks: encoder, recovery (decoder), generator, supervisor, and discriminator. We extend the base model to incorporate metadata. Each network uses a three-layer GRU~\cite{cho2014properties} with fully connected output heads.

For the diffusion backbones, we use a linear noise schedule with 200 noising and denoising steps, using $\alpha_1 = 0.9999$ and $\alpha_T = 0.98$. The guidance strength $\eta$ is fixed at 0.1 for all datasets and task configurations. By default, we report \algo's results using a stride of 8, as it achieved the best overall performance. We trained the diffusion backbones using the MSE loss between the estimated and actual noise, and we optimised TimeGAN using the L1-regularised reconstruction loss between the generated and ground-truth signals. We train all models for 300 epochs using a mini-batch size of 1024 and a learning rate of $10^{-4}$. For consistency, we divided all the datasets using sliding windows of 32 timesteps (stride 1) for training, and tunable stride lengths at inference.

\begin{table}[th]
\centering
\caption{{Average MSE values ($\downarrow$) for different datasets and tasks (\textit{R}/\textit{I}/\textit{B}). Standard deviation in subscript; \textbf{best} (bold) and \underline{second-best} scores marked.}}
\label{tab:mse_table}
\resizebox{\textwidth}{!}{
\begin{tabular}{l llllll}
\Xhline{1.5pt}
\textbf{} & \textbf{TimeGAN}\phantom{ddd} & \textbf{TimeWeaver} & \textbf{TSDiff}\phantom{dddd} & \textbf{SSSD}\phantom{ddddd} & \makecell[c]{\textbf{Time-}\\\textbf{AutoDiff}} & \textbf{\algo} \\
\midrule
\rowcolor{lightgray}
\phantom{AT} (\bemph{R}) & $ 8.159_{.115}$ & $ \underline{0.295_{.040}}$ & $ 1.223_{.007}$ & $ 1.329_{.004}$ & $ 10.406_{.433}$ & $ \mathbf{0.152_{.006}}$ \\
\rowcolor{lightgray}
AT (\bemph{I}) & $ 8.373_{.062}$ & $ \underline{1.155_{.335}}$ & $ 1.480_{.003}$ & $ 1.490_{.011}$ & $ 6.200_{.407}$ & $ \mathbf{0.246_{.009}}$ \\
\rowcolor{lightgray}
\phantom{AT} (\bemph{B}) & $ 9.717_{.125}$ & $ 2.253_{.006}$ & $ 1.567_{.032}$ & $ \underline{0.652_{.030}}$ & $ 4.021_{.144}$ & $ \mathbf{0.140_{.004}}$ \\
\phantom{MT} (\bemph{R}) & $ 8.887_{.072}$ & $ \underline{0.533_{.018}}$ & $ 0.964_{.009}$ & $ 1.009_{.010}$ & $ 3.957_{.040}$ & $ \mathbf{0.512_{.009}}$ \\
{MT} (\bemph{I}) & $ 9.194_{.175}$ & $ \underline{0.603_{.075}}$ & $ 0.768_{.027}$ & $ 0.701_{.061}$ & $ 1.667_{.088}$ & $ \mathbf{0.396_{.019}}$ \\
\phantom{MT} (\bemph{B}) & $ 7.817_{.222}$ & $ 0.874_{.042}$ & $ 0.200_{.010}$ & $ \underline{0.128_{.005}}$ & $ 0.702_{.033}$ & $ \mathbf{0.111_{.008}}$ \\
\rowcolor{lightgray}
\phantom{BQ} (\bemph{R}) & $ 3.148_{.074}$ & $ 1.529_{.022}$ & $ 1.763_{.017}$ & $ \mathbf{1.330_{.022}}$ & $ 16.901_{.122}$ & $ \underline{1.481_{.019}}$ \\
\rowcolor{lightgray}
BQ (\bemph{I}) & $ 3.552_{.177}$ & $ 1.067_{.021}$ & $ 1.422_{.025}$ & $ \underline{1.055_{.016}}$ & $ 23.695_{.427}$ & $ \mathbf{0.981_{.015}}$ \\
\rowcolor{lightgray}
\phantom{BQ} (\bemph{B}) & $ 2.170_{.039}$ & $ 1.595_{.045}$ & $ 0.162_{.007}$ & $ \underline{0.107_{.002}}$ & $ 1.085_{.019}$ & $ \mathbf{0.099_{.009}}$ \\
\phantom{RS} (\bemph{R}) & $ 4.052_{.003}$ & $ 0.645_{.014}$ & $ 0.893_{.021}$ & $ \underline{0.640_{.005}}$ & $ 4.680_{.066}$ & $ \mathbf{0.598_{.006}}$ \\
RS (\bemph{I}) & $ 4.065_{.015}$ & $ 0.588_{.043}$ & $ 0.827_{.032}$ & $ \underline{0.569_{.014}}$ & $ 2.839_{.081}$ & $ \mathbf{0.537_{.023}}$ \\
\phantom{RS} (\bemph{B}) & $ 2.731_{.020}$ & $ 0.965_{.059}$ & $ 0.458_{.011}$ & $ \underline{0.440_{.005}}$ & $ 1.019_{.006}$ & $ \mathbf{0.149_{.002}}$ \\
\rowcolor{lightgray}
\phantom{PE} (\bemph{R}) & $ 6.768_{.007}$ & $ \underline{1.074_{.027}}$ & $ 1.667_{.005}$ & $ 2.496_{.008}$ & $ 4.127_{.031}$ & $ \mathbf{1.006_{.013}}$ \\
\rowcolor{lightgray}
PE (\bemph{I}) & $ 6.138_{.102}$ & $ \underline{0.961_{.114}}$ & $ 1.179_{.061}$ & $ 1.121_{.055}$ & $ 9.279_{.422}$ & $ \mathbf{0.587_{.128}}$ \\
\rowcolor{lightgray}
\phantom{PE} (\bemph{B}) & $ 4.092_{.009}$ & $ 1.387_{.029}$ & $ 0.469_{.002}$ & $ \underline{0.216_{.001}}$ & $ 1.378_{.003}$ & $ \mathbf{0.165_{.003}}$ \\

\Xhline{1.5pt}
\end{tabular}
}
\end{table}

\begin{table}[th]
\centering
\caption{{Average ACD values ($\downarrow$) for each dataset and task (\textit{R}/\textit{I}/\textit{B}). Standard deviation in subscript; \textbf{best} (bold) and \underline{second-best} scores marked.)}}
\label{tab:acd_table}
\resizebox{\textwidth}{!}{
\begin{tabular}{lllllll}
\Xhline{1.5pt}
\textbf{} & \textbf{TimeGAN}\phantom{ddd} & \textbf{TimeWeaver} & \textbf{TSDiff}\phantom{dddd} &\textbf{SSSD}\phantom{ddddd} & \makecell[c]{\textbf{Time-}\\\textbf{AutoDiff}} & \textbf{\algo} \\
\midrule
\rowcolor{lightgray}
\phantom{AT} (\bemph{R}) & $ 0.259_{.015}$ & $ \underline{0.024_{.003}}$ & $ 0.090_{.003}$ & $ 0.131_{.003}$ & $ 0.151_{.004}$ & $ \mathbf{0.018_{.002}}$ \\
\rowcolor{lightgray}
AT (\bemph{I}) & $ 0.253_{.006}$ & $ \underline{0.093_{.021}}$ & $ 0.256_{.002}$ & $ 0.129_{.003}$ & $ 0.226_{.010}$ & $ \mathbf{0.038_{.001}}$ \\
\rowcolor{lightgray}
\phantom{AT} (\bemph{B}) & $ 0.255_{.024}$ & $ 0.126_{.002}$ & $ 0.090_{.001}$ & $ \underline{0.057_{.001}}$ & $ 0.111_{.005}$ & $ \mathbf{0.022_{.000}}$ \\
\phantom{MT} (\bemph{R}) & $ 0.467_{.004}$ & $ \underline{0.191_{.005}}$ & $ 0.308_{.006}$ & $ 0.332_{.002}$ & $ 0.343_{.006}$ & $ \mathbf{0.142_{.007}}$ \\
{MT} (\bemph{I}) & $ 0.437_{.008}$ & $ 0.208_{.025}$ & $ \underline{0.159_{.008}}$ & $ 0.181_{.012}$ & $ 0.238_{.007}$ & $ \mathbf{0.134_{.010}}$ \\
\phantom{MT} (\bemph{B}) & $ 0.403_{.024}$ & $ 0.234_{.038}$ & $ 0.057_{.001}$ & $ \underline{0.045_{.000}}$ & $ 0.063_{.002}$ & $ \mathbf{0.042_{.001}}$ \\
\rowcolor{lightgray}
\phantom{BQ} (\bemph{R}) & $ 0.191_{.002}$ & $ \underline{0.154_{.005}}$ & $ 0.284_{.002}$ & $ 0.303_{.004}$ & $ 0.285_{.004}$ & $ \mathbf{0.127_{.003}}$ \\
\rowcolor{lightgray}
BQ (\bemph{I}) & $ 0.231_{.006}$ & $ \mathbf{0.104_{.003}}$ & $ 0.171_{.006}$ & $ 0.219_{.004}$ & $ 0.321_{.001}$ & $ \underline{0.116_{.010}}$ \\
\rowcolor{lightgray}
\phantom{BQ} (\bemph{B}) & $ 0.223_{.005}$ & $ 0.265_{.021}$ & $ 0.025_{.001}$ & $ \mathbf{0.018_{.000}}$ & $ 0.096_{.001}$ & $ \mathbf{0.020_{.001}}$ \\
\phantom{RS} (\bemph{R}) & $ 0.289_{.002}$ & $ 0.150_{.001}$ & $ \mathbf{0.131_{.002}}$ & $ \underline{0.143_{.001}}$ & $ 0.218_{.005}$ & $ 0.162_{.001}$ \\
RS (\bemph{I}) & $ 0.289_{.005}$ & $ 0.181_{.004}$ & $ \underline{0.176_{.005}}$ & $ \mathbf{0.137_{.006}}$ & $ 0.194_{.010}$ & $ 0.183_{.007}$ \\
\phantom{RS} (\bemph{B}) & $ 0.311_{.004}$ & $ 0.241_{.005}$ & $ \underline{0.126_{.006}}$ & $ 0.128_{.005}$ & $ 0.213_{.000}$ & $ \mathbf{0.051_{.001}}$ \\
\rowcolor{lightgray}
\phantom{PE} (\bemph{R}) & $ 0.410_{.000}$ & $ \underline{0.321_{.003}}$ & $ 0.388_{.001}$ & $ 0.422_{.001}$ & $ 0.424_{.001}$ & $ \mathbf{0.294_{.005}}$ \\
\rowcolor{lightgray}
PE (\bemph{I}) & $ 0.382_{.002}$ & $ 0.258_{.020}$ & $ 0.309_{.011}$ & $ \underline{0.252_{.005}}$ & $ 0.414_{.013}$ & $ \mathbf{0.178_{.032}}$ \\
\rowcolor{lightgray}
\phantom{PE} (\bemph{B}) & $ 0.417_{.001}$ & $ 0.353_{.024}$ & $ \underline{0.081_{.001}}$ & $ \mathbf{0.044_{.000}}$ & $ 0.333_{.000}$ & $ \underline{0.081_{.007}}$ \\
\Xhline{1.5pt}
\end{tabular}}
\end{table}

\begin{table}[th]
\centering
\caption{{Average $x$-Corr values ($\downarrow$) for each dataset and task (\textit{R}/\textit{I}/\textit{B}). Standard deviation in subscript; \textbf{best} (bold) and \underline{second-best} scores marked.}}
\label{tab:xcorr_pivoted}

\resizebox{\textwidth}{!}{
\begin{tabular}{lllllll}
\Xhline{1.5pt}
\textbf{} & \textbf{TimeGAN}\phantom{ddd} & \textbf{TimeWeaver} & \textbf{TSDiff}\phantom{dddd} &\textbf{SSSD}\phantom{ddddd} & \makecell[c]{\textbf{Time-}\\\textbf{AutoDiff}} & \textbf{\algo} \\
\midrule
\rowcolor{lightgray}
\phantom{MT} (\bemph{R}) & $ 0.558_{.002}$ & $ \mathbf{0.102_{.006}}$ & $ 0.120_{.003}$ & $ 0.204_{.004}$ & $ 0.150_{.005}$ & $ \underline{0.104_{.003}}$ \\
\rowcolor{lightgray}
{MT} (\bemph{I}) & $ 0.587_{.017}$ & $ 0.203_{.024}$ & $ \underline{0.180_{.013}}$ & $ \mathbf{0.179_{.026}}$ & $ 0.205_{.020}$ & $ 0.206_{.013}$ \\
\rowcolor{lightgray}
\phantom{MT} (\bemph{B}) & $ 0.527_{.015}$ & $ 0.105_{.015}$ & $ \mathbf{0.062_{.005}}$ & $ 0.091_{.005}$ & $ 0.089_{.011}$ & $ \underline{0.074_{.005}}$ \\
\phantom{BQ} (\bemph{R}) & $ 0.191_{.004}$ & $ \mathbf{0.156_{.002}}$ & $ \underline{0.178_{.003}}$ & $ 0.284_{.003}$ & $ 0.341_{.002}$ & $ \mathbf{0.156_{.008}}$ \\
BQ (\bemph{I}) & $ 0.232_{.006}$ & $ \underline{0.191_{.015}}$ & $ \mathbf{0.187_{.003}}$ & $ 0.200_{.004}$ & $ 0.371_{.002}$ & $ \mathbf{0.187_{.015}}$ \\
\phantom{BQ} (\bemph{B}) & $ 0.181_{.006}$ & $ 0.285_{.032}$ & $ 0.047_{.002}$ & $ \mathbf{0.019_{.000}}$ & $ 0.230_{.002}$ & $ \underline{0.038_{.003}}$ \\
\rowcolor{lightgray}
\phantom{RS} (\bemph{R}) & $ 0.703_{.005}$ & $ \underline{0.030_{.001}}$ & $ 0.046_{.001}$ & $ 0.026_{.000}$ & $ \mathbf{0.004_{.000}}$ & $ 0.031_{.001}$ \\
\rowcolor{lightgray}
RS (\bemph{I}) & $ 0.741_{.015}$ & $ \underline{0.006_{.001}}$ & $ \mathbf{0.004_{.002}}$ & $ 0.009_{.001}$ & $ \mathbf{0.004_{.002}}$ & $ 0.009_{.001}$ \\
\rowcolor{lightgray}
\phantom{RS} (\bemph{B}) & $ 0.829_{.018}$ & $ 0.127_{.008}$ & $ 0.075_{.003}$ & $ 0.073_{.002}$ & $ \underline{0.047_{.001}}$ & $ \mathbf{0.011_{.001}}$ \\
\phantom{PE} (\bemph{R}) & $ 0.412_{.001}$ & $ \mathbf{0.145_{.005}}$ & $ 0.298_{.003}$ & $ \underline{0.195_{.001}}$ & $ 0.255_{.002}$ & $ 0.228_{.006}$ \\
PE (\bemph{I}) & $ 0.535_{.007}$ & $ \mathbf{0.175_{.043}}$ & $ 0.321_{.021}$ & $ 0.344_{.009}$ & $ 0.293_{.023}$ & $ \underline{0.232_{.017}}$ \\
\phantom{PE} (\bemph{B}) & $ 0.366_{.001}$ & $ 0.249_{.025}$ & $ 0.099_{.001}$ & $ \mathbf{0.031_{.000}}$ & $ 0.160_{.001}$ & $ \underline{0.056_{.003}}$ \\
\Xhline{1.5pt}
\end{tabular}}
\end{table}

\subsection{Comparison with SOTA Generators (RQ1)}
\label{ssec:baseline}
\autoref{tab:mse_table} shows the average MSE between the generated and ground truth signals over five trials. We use multiple metrics to assess performance: MSE for imputation accuracy, and quality metrics ACD and $x$-Corr for generative tasks.

\begin{table*}[htb]
\centering
\caption{Autoregressive (\algoar) versus parallel generation across various strides (1, 8, 16, 32), datasets and tasks. The best scores are marked in \textbf{bold}.} 
\label{tab:paraustr}
\resizebox{\textwidth}{!}{
\begin{tabular}{lllllllllll}
    \bottomrule
    \Xhline{1.5pt}
    && \multicolumn{3}{c}{\bemph{R}} & \multicolumn{3}{c}{\bemph{I}} & \multicolumn{3}{c}{\bemph{B}} \\
    \cmidrule(lr){3-5} \cmidrule(lr){6-8} \cmidrule(lr){9-11}
    &\textbf{Method} & \textbf{Avg. MSE $\downarrow$} & \textbf{\#Calls $\downarrow$} & \textbf{Avg. Time $\downarrow$} & \textbf{Avg. MSE $\downarrow$} & \textbf{\#Calls $\downarrow$} & \textbf{Avg. Time $\downarrow$} & \textbf{Avg. MSE $\downarrow$} & \textbf{\#Calls $\downarrow$} & \textbf{Avg. Time $\downarrow$} \\
    \cmidrule(lr){2-11}
    \multirow{7}{*}{\rotatebox{90}{AT}}&\algoar-8 & $0.214_{0.033}$ & $30800$ & $298.644_{2.018}$ & $\mathbf{0.239_{0.004}}$ & $5200$ & $50.279_{0.449}$ & $0.145_{0.003}$ & $30800$ & $297.488_{1.731}$\\
 &\algoar-16 & $0.177_{0.022}$ & $15400$ & $148.225_{1.224}$ & $0.256_{0.039}$ & $3000$ & $28.859_{0.546}$ & $0.159_{0.031}$ & $15400$ & $149.243_{0.529}$\\
 &\algoar-32 & $0.290_{0.041}$ & $7600$ & $73.919_{0.901}$ & $0.350_{0.094}$ & $1800$ & $17.511_{0.563}$ & $\mathbf{0.138_{0.002}}$ & $7600$ & $73.957_{0.628}$\\
 &\algo-1 & $0.167_{0.008}$ & $400$ & $17.838_{0.650}$ & $0.288_{0.045}$ & $400$ & $17.784_{0.842}$ & $0.152_{0.003}$ & $400$ & $17.518_{0.572}$\\
 &\algo-8 & $\mathbf{0.152_{0.006}}$ & $200$ & $\mathbf{5.258_{0.579}}$ & $0.246_{0.009}$ & $200$ & $\mathbf{5.447_{0.744}}$ & $0.140_{0.004}$ & $200$ & $\mathbf{5.283_{0.623}}$\\
 &\algo-16 & $0.166_{0.012}$ & $200$ & $5.642_{0.640}$ & $0.249_{0.005}$ & $200$ & $5.819_{0.615}$ & $0.140_{0.004}$ & $200$ & $5.336_{0.748}$\\
 &\algo-32 & $0.304_{0.062}$ & $200$ & $5.549_{0.588}$ & $0.715_{0.147}$ & $200$ & $5.582_{0.588}$ & $0.194_{0.065}$ & $200$ & $5.368_{0.645}$\\
 \cdashline{2-11}
 &TimeGAN-32 & $8.159_{0.115}$ & $1$ & $0.020_{0.036}$ & $8.373_{0.062}$ & $1$ & $0.019_{0.037}$ & $9.717_{0.125}$ & $1$ & $0.096_{0.010}$\\
    \bottomrule
    \multirow{7}{*}{\rotatebox{90}{MT}}&\algoar-8 & $0.827_{0.111}$ & $198600$ & $1937.021_{6.488}$ & $0.367_{0.017}$ & $8400$ & $83.225_{1.212}$ & $0.130_{0.009}$ & $56600$ & $546.098_{1.696}$\\
 &\algoar-16 & $\mathbf{0.504_{0.042}}$ & $99200$ & $971.742_{3.573}$ & $\mathbf{0.350_{0.068}}$ & $5200$ & $51.712_{0.523}$ & $0.119_{0.014}$ & $55600$ & $540.363_{2.691}$\\
 &\algoar-32 & $0.548_{0.007}$ & $49600$ & $488.252_{1.868}$ & $0.369_{0.025}$ & $3400$ & $33.458_{0.541}$ & $0.125_{0.008}$ & $46600$ & $451.989_{1.995}$\\
 &\algo-1 & $0.522_{0.011}$ & $1600$ & $94.320_{0.643}$ & $0.390_{0.061}$ & $1600$ & $94.384_{1.069}$ & $\mathbf{0.109_{0.008}}$ & $1600$ & $94.499_{1.471}$\\
 &\algo-8 & $0.512_{0.009}$ & $200$ & $11.262_{0.575}$ & $0.396_{0.019}$ & $200$ & $11.497_{0.533}$ & $0.111_{0.008}$ & $200$ & $11.164_{0.533}$\\
 &\algo-16 & $0.506_{0.012}$ & $200$ & $6.848_{0.509}$ & $0.378_{0.055}$ & $200$ & $7.059_{0.479}$ & $0.115_{0.007}$ & $200$ & $6.822_{0.542}$\\
 &\algo-32 & $0.547_{0.016}$ & $200$ & $\mathbf{5.827_{0.499}}$ & $0.410_{0.038}$ & $200$ & $\mathbf{5.901_{0.629}}$ & $0.121_{0.011}$ & $200$ & $\mathbf{5.709_{0.637}}$\\
 \cdashline{2-11}
 &TimeGAN-32 & $8.887_{0.072}$ & $1$ & $0.027_{0.050}$ & $9.194_{0.175}$ & $1$ & $0.019_{0.036}$ & $7.817_{0.222}$ & $1$ & $0.020_{0.038}$\\
     \bottomrule
     \multirow{7}{*}{\rotatebox{90}{BQ}} &\algoar-8 & $1.764_{0.526}$ & $212400$ & $2070.688_{9.150}$ & $\mathbf{0.752_{0.014}}$ & $100800$ & $977.191_{4.176}$ & ${0.096_{0.005}}$ & $70800$ & $686.126_{4.232}$\\
 &\algoar-16 & $2.650_{0.098}$ & $106200$ & $1033.736_{3.506}$ & $1.132_{0.139}$ & $51000$ & $503.286_{2.742}$ & $0.104_{0.006}$ & $70800$ & $685.728_{1.410}$\\
 &\algoar-32 & $1.511_{0.010}$ & $53000$ & $510.891_{1.523}$ & $1.034_{0.013}$ & $26000$ & $255.775_{1.753}$ & $\mathbf{0.095_{0.008}}$ & $53000$ & $523.066_{1.998}$\\
 
 &\algo-1 & $1.523_{0.022}$ & $1800$ & $101.472_{0.605}$ & $1.043_{0.015}$ & $1800$ & $101.482_{0.607}$ & $0.104_{0.008}$ & $1800$ & $101.603_{0.719}$\\
 
 &\algo-8 & $\mathbf{1.481_{0.019}}$ & $400$ & $17.776_{0.634}$ & $0.981_{0.015}$ & $400$ & $17.564_{0.733}$ & $0.099_{0.009}$ & $400$ & $17.455_{0.762}$\\
 &\algo-16 & $1.488_{0.025}$ & $200$ & $7.294_{0.518}$ & $0.999_{0.011}$ & $200$ & $7.319_{0.609}$ & $0.101_{0.010}$ & $200$ & $7.377_{0.603}$\\
 
 &\algo-32 & $1.530_{0.014}$ & $200$ & $\mathbf{5.931_{0.523}}$ & $1.051_{0.015}$ & $200$ & $\mathbf{5.987_{0.506}}$ & $0.111_{0.011}$ & $200$ & $\mathbf{5.999_{0.768}}$\\
 \cdashline{2-11}
 &TimeGAN-32 & $3.148_{0.074}$ & $1$ & $0.026_{0.047}$ & $3.552_{0.177}$ & $1$ & $0.029_{0.055}$ & $2.170_{0.039}$ & $1$ & $0.019_{0.037}$\\
     \bottomrule
     \multirow{7}{*}{\rotatebox{90}{RS}} &\algoar-8 & $0.647_{0.007}$ & $43800$ & $425.083_{1.988}$ & $0.606_{0.022}$ & $6600$ & $64.161_{0.769}$ & $0.173_{0.004}$ & $34800$ & $338.959_{1.832}$\\
 &\algoar-16 & $0.609_{0.011}$ & $21800$ & $211.279_{0.999}$ & $0.581_{0.045}$ & $3400$ & $32.882_{0.559}$ & $0.156_{0.001}$ & $21800$ & $211.105_{0.409}$\\
 &\algoar-32 & $0.642_{0.007}$ & $10800$ & $102.649_{0.884}$ & $0.643_{0.028}$ & $1800$ & $17.595_{0.532}$ & $0.155_{0.001}$ & $10800$ & $105.553_{0.574}$\\
 
 &\algo-1 & $\mathbf{0.593_{0.008}}$ & $400$ & $20.089_{0.579}$ & $0.542_{0.022}$ & $400$ & $20.110_{0.536}$ & $0.165_{0.002}$ & $400$ & $20.092_{0.527}$\\
 
 &\algo-8 & $0.598_{0.006}$ & $200$ & $5.279_{0.581}$ & $\mathbf{0.537_{0.023}}$ & $200$ & $5.210_{0.556}$ & $\mathbf{0.149_{0.002}}$ & $200$ & $5.330_{0.502}$\\
 
 &\algo-16 & $0.597_{0.012}$ & $200$ & $\mathbf{5.129_{0.648}}$ & $0.553_{0.040}$ & $200$ & $\mathbf{4.956_{0.432}}$ & $0.150_{0.001}$ & $200$ & $\mathbf{5.184_{0.508}}$\\
 
 &\algo-32 & $0.618_{0.009}$ & $200$ & $5.798_{0.524}$ & $0.605_{0.029}$ & $200$ & $5.306_{0.889}$ & $0.158_{0.002}$ & $200$ & $5.673_{0.536}$\\
  \cdashline{2-11}
 &TimeGAN-32 & $4.052_{0.003}$ & $1$ & $0.030_{0.057}$ & $4.065_{0.015}$ & $1$ & $0.019_{0.037}$ & $2.731_{0.020}$ & $1$ & $0.021_{0.040}$\\
     \bottomrule
     \multirow{7}{*}{\rotatebox{90}{PE}} &\algoar-8 & $2.417_{0.020}$ & $320400$ & $3139.356_{10.071}$ & $\mathbf{0.232_{0.016}}$ & $10800$ & $105.714_{0.831}$ & $0.184_{0.003}$ & $320400$ & $3153.572_{4.575}$\\
 &\algoar-16 & $1.102_{0.027}$ & $160200$ & $1566.418_{6.249}$ & $0.232_{0.038}$ & $6000$ & $58.881_{0.689}$ & $0.171_{0.002}$ & $160200$ & $1557.470_{3.514}$\\
 &\algoar-32 & $1.071_{0.024}$ & $80000$ & $778.270_{2.683}$ & $0.966_{0.074}$ & $3600$ & $34.829_{0.449}$ & $0.175_{0.003}$ & $80000$ & $771.878_{3.205}$\\
 
 &\algo-1 & $1.046_{0.016}$ & $2600$ & $149.726_{0.569}$ & $0.718_{0.126}$ & $2600$ & $149.801_{0.651}$ & $0.171_{0.003}$ & $2600$ & $149.827_{0.568}$\\
 
 &\algo-8 & $\mathbf{1.006_{0.013}}$ & $400$ & $19.556_{0.609}$ & $0.587_{0.128}$ & $400$ & $19.532_{0.555}$ & $\mathbf{0.165_{0.003}}$ & $400$ & $19.485_{0.553}$\\
 
 &\algo-16 & $1.017_{0.017}$ & $200$ & $9.409_{0.585}$ & $0.606_{0.087}$ & $200$ & $9.376_{0.552}$ & $0.165_{0.005}$ & $200$ & $9.439_{0.536}$\\
 
 &\algo-32 & $1.073_{0.022}$ & $200$ & $\mathbf{6.392_{0.467}}$ & $0.688_{0.053}$ & $200$ & $\mathbf{6.445_{0.534}}$ & $0.183_{0.004}$ & $200$ & $\mathbf{6.471_{0.605}}$\\
 \cdashline{2-11}
 &TimeGAN-32 & $6.768_{0.007}$ & $1$ & $0.029_{0.054}$ & $6.138_{0.102}$ & $1$ & $0.020_{0.038}$ & $4.092_{0.009}$ & $1$ & $0.020_{0.037}$\\
     \bottomrule
     \Xhline{1.5pt}
\end{tabular}
}
\end{table*}

We observe that diffusion methods outperform TimeGAN, in line with earlier findings~\cite{dhariwal2021diffusion,shankar2024silofuse}. MSEs generally decrease from \emph{R} tasks to \emph{I} and \emph{B} tasks, since \emph{R} tasks have a single contiguous missing block. In contrast, \emph{I} and \emph{B} tasks require generating several smaller blocks separated by observed signals. {\color{black}When computing the MSE ratio by dividing the best-performing baseline's MSE with that of \ algo's for each task and dataset, the average ratio is \textcolor{black}{\textbf{1.81x}}, highlighting \ algo's significant advantage over SOTA baselines.}  

{\color{black}
Among the baselines, SSSD, TSDiff and TimeWeaver generally perform the best. TimeWeaver performs well on \textit{R} (coarse-grained) tasks, but underperforms on fine-grained tasks since it ignores the local cues from neighbouring signal observations. In contrast, TSDiff and SSSD do better on fine-grained tasks since they also condition on observed signals. SSSD, trained by randomly masking signal rows, likely encounters many patterns where masked entries are neighboured by observations, possibly explaining its stronger performance on fine-grained tasks. However, this bias towards the masking patterns seen during training could explain its underperformance on coarser-grained tasks. TimeAutoDiff performs poorly across the board, likely due to its sensitivity to the guidance strength and incompatibilities from using two models.}

For synthesis tasks, we assess methods based on their ability to preserve temporal patterns and inter-feature dependencies, using ACD and $x$-Corr scores (\autoref{tab:acd_table} and \autoref{tab:xcorr_pivoted}). The AT dataset is excluded from \autoref{tab:xcorr_pivoted} since it is univariate and does not have inter-feature dependencies. We see that \algo often performs the best or second-best. Exceptions are likely due to the conditioning being biased towards minimising an MSE-like loss, which may not always yield the best results for ACD and $x$-Corr. Nevertheless, its performance is still competitive. 

Again, while TimeWeaver performs well at root-level tasks, it underperforms on fine-grained tasks. In contrast, conditioning on the signals observed enables TSDiff, SSSD, and \algo to generate more accurate values. TimeGAN consistently underperforms across all metrics, further highlighting the advantage of diffusion-based models for generating high-quality time series data.

\subsection{Ablation: Parallel vs. Autoregressive (RQ2)}
\label{ssec:parallel}
 \autoref{tab:paraustr} shows the MSE, average runtime (seconds) and the total number of denoiser calls of the parallel (\algo) and autoregressive (\algoar) implementations under varying stride lengths. We average all results over five trials, using a window size of 32 consistent across all variants, with 200 denoising steps and a {\color{black}mini-batch} size of 1024. {\color{black} We also report TimeGAN results to contextualise the tradeoff between generation quality and speed against a fast generative baseline.}

{\color{black}Overall, the MSE scores between the parallel (\algo) and autoregressive (\algoar) variants are comparable, with the parallel version generally doing better. This result indicates that parallel synthesis with stitching maintains similar coherence to sequential generation. While both versions share the same model and sliding window setup, they differ in how they apply conditioning over time. In \algoar, each generated window is used as context for the next one, so early errors can propagate unchecked, degrading quality for the subsequent windows. In contrast, \algo imposes conditions ``softly'' via the conditional loss, iteratively refining all windows together during denoising. This joint optimisation improves robustness to early mistakes, giving \algo a slight edge over \algoar.} 

The parallel approach also has a much faster runtime. For example, in task \emph{R} in the PE dataset, \algo-16 (stride 16) achieves an \textbf{average speedup of} $\mathbf{166.48}$ (i.e., $\frac{1566.418}{9.409}$) over its sequential counterpart \algoar-16. It also achieves a {lower MSE} and reduces the function calls from \textbf{160200} to just \textbf{200}. The runtime generally decreases with larger strides, resulting in fewer windows, thus resulting in fewer denoiser calls. However, for shorter sequences like AT, where nearly all windows fit within one {\color{black}mini-batch} (1024), the stride length has little impact on runtime. 

The effect of stitching becomes evident with \algo-32, which follows a \emph{divide-and-conquer} approach without overlaps (and hence no stitching). It typically performs worse than variants with smaller strides, where stitching allows more contextual information to guide generation. However, excessive overlap can also propagate errors from earlier windows, so using smaller strides does not always yield better accuracy.

The speedup does not reach the theoretical limit of the {\color{black}mini-batch} size $b$ in practice (see Sec.~\ref{subsubsec:speed}), possibly due to the under-utilisation of hardware and other overheads. For example, consider the MT dataset, with a sequence length of 7949 timesteps (see \autoref{tab:datasets}). For window size 32 and stride 8, the number of windows is roughly $(7949 - 32) / 8 \approx 990$. As this is less than the {\color{black}mini-batch} size of 1024, the autoregressive model makes roughly\footnote{The exact total is 198600 due to a minor implementation detail where additional timesteps are added for conditional context.} $990 \times 200 = 198000$ denoiser calls. This is lower than the worst-case with $1024 \times 200 = 204800$ calls, so we have a lower speedup.

{\color{black}
When analysing the speed–quality tradeoff, diffusion models consistently have lower MSEs than TimeGAN despite slower speeds. This difference stems from the core design: while diffusion iteratively corrects errors at each step, GANs generate using a single step, which is unstable.

For \algo-32 (generally the fastest variant), we see the average time \textit{per denoiser call} ranges from $\sim$0.026-0.032 seconds, which is close to TimeGAN ($\sim$0.019-0.030 seconds). Thus, the overhead of diffusion stems not from individual operations, but from the larger number of denoising steps. We can achieve further efficiency improvements by reducing the number of steps using techniques like model distillation~\cite{salimans2022progressive, Meng_2023_CVPR}. These methods can be applied post-hoc on top of our design to further improve sampling efficiency.}

\subsection{Ablation: Effect of Stitching (RQ3)}

{
\autoref{tab:conditioning} evaluates the imputation accuracy (MSE) using five trials, of three conditioning strategies: (1) \algo with RePaint-based conditioning~\cite{lugmayr2022repaint} and enforced overlap-coherence; (2) \algo with only self-guidance on observations without stitching; and 
(3) \algo with self-guidance and stitching. 
Results show that using stitching achieves the best performance in nearly all cases, while relying only on self-guidance performs the worst. RePaint is competitive, but generally loses to ours. Overall, the stitch loss improves performance by up to \textbf{35.43\%} compared to just self-guidance ({AT, \textit{I} task}) and up to \textbf{28.36\%} over RePaint ({RS, \textit{B} task}). This result directly aligns with Proposition 3.1 (Sec.~\ref{subsec:theory}), which explains how using self-guidance and stitching increases the likelihood of generating realistic samples.
}
\begin{table}[htb]
\centering
\caption{\centering{Comparing conditioning methods on imputation accuracy (MSE).}}
\label{tab:conditioning}

\begin{tabular}{lcccccc}
\Xhline{1.5pt}
\textbf{} & \textbf{RePaint} & \textbf{Self-guidance} & \makecell[c]{\textbf{Self-guidance} \\ \textbf{+ Stitching}} \\
\midrule
\rowcolor{lightgray}
\phantom{AT} (\bemph{R}) & $ 0.167_{.012}$ & $ 0.222_{.030}$ & $ \mathbf{0.152_{.006}}$ \\
\rowcolor{lightgray}
AT (\bemph{I}) & $ 0.278_{.046}$ & $ 0.381_{.054}$ & $ \mathbf{0.246_{.009}}$ \\
\rowcolor{lightgray}
\phantom{AT} (\bemph{B}) & $ 0.153_{.007}$ & $ 0.178_{.024}$ & $ \mathbf{0.140_{.004}}$ \\
\phantom{MT} (\bemph{R}) & $ \mathbf{0.506_{.009}}$ & $ 0.544_{.010}$ & $ 0.512_{.009}$ \\
MT (\bemph{I}) & $ 0.439_{.037}$ & $ 0.470_{.048}$ & $ \mathbf{0.396_{.019}}$ \\
\phantom{MT} (\bemph{B}) & $ 0.140_{.011}$ & $ 0.127_{.007}$ & $ \mathbf{0.111_{.008}}$ \\
\rowcolor{lightgray}
\phantom{BQ} (\bemph{R}) & $ 1.525_{.017}$ & $ 1.541_{.018}$ & $ \mathbf{1.481_{.019}}$ \\
\rowcolor{lightgray}
BQ (\bemph{I}) & $ 1.027_{.015}$ & $ 1.050_{.017}$ & $ \mathbf{0.981_{.015}}$ \\
\rowcolor{lightgray}
\phantom{BQ} (\bemph{B}) & $ 0.111_{.005}$ & $ 0.102_{.010}$ & $ \mathbf{0.099_{.009}}$ \\
\phantom{RS} (\bemph{R}) & $ 0.616_{.005}$ & $ 0.615_{.005}$ & $ \mathbf{0.598_{.006}}$ \\
{RS} (\bemph{I}) & $ 0.570_{.020}$ & $ 0.561_{.022}$ & $ \mathbf{0.537_{.023}}$ \\
\phantom{RS} (\bemph{B}) & $ 0.208_{.005}$ & $ 0.168_{.003}$ & $ \mathbf{0.149_{.002}}$ \\
\rowcolor{lightgray}
\phantom{PE} (\bemph{R}) & $ 1.016_{.013}$ & $ 1.063_{.013}$ & $ \mathbf{1.006_{.013}}$ \\
\rowcolor{lightgray}
PE (\bemph{I}) & $ \mathbf{0.511_{.102}}$ & $ 0.847_{.134}$ & $ 0.587_{.128}$ \\
\rowcolor{lightgray}
\phantom{PE} (\bemph{B}) & $ 0.177_{.002}$ & $ 0.178_{.003}$ & $ \mathbf{0.165_{.003}}$ \\
\Xhline{1.5pt}
\end{tabular}
\end{table}

{\color{black}\subsection{Choice of Stitch Loss Function (RQ4)}

\autoref{tab:stitchloss} compares the imputation accuracy, measured by MSE, of four stitch loss formulations for time series: Mean Absolute Error (MAE), Cosine similarity, Pearson correlation loss \cite{berthold2016clustering}, and our MSE-based loss. These losses include distance-based objectives (MSE, MAE) and shape-based criteria (cosine similarity, Pearson correlation). Each one promotes alignment differently: Pearson loss promotes linear dependence between the overlaps along the time axis; cosine similarity pushes the waveform shapes to match; and MAE is more robust towards outliers than MSE.

We observe that distance-based losses (MSE, MAE) generally outperform the shape-based objectives. This is because a zero stitch loss for distance-based objectives directly forces values across the overlaps to match. In contrast, shape-similarity losses can become zero even when overlaps differ in absolute value. Although we adopt MSE for stitching due to its simplicity and ease of optimisation, these results indicate that using alternatives, such as MAE, is still viable for task-specific flexibility.}
\begin{table}[htb]
\centering
\caption{\centering{{\color{black}Comparison of stitch loss functions (MSE$\downarrow$), for various datasets and tasks (R/I/B).}}}
\label{tab:stitchloss}

\begin{tabular}{lcccccccc}
\Xhline{1.5pt}
\textbf{} & \textbf{MSE} & \textbf{MAE} & \textbf{Cosine} & \textbf{Correlation}  \\
\midrule
\rowcolor{lightgray}
\phantom{AT} (\bemph{R}) & $ \mathbf{0.152_{.006}}$ & $ 0.157_{.006}$ & $ 0.182_{.021}$ & $ 0.182_{.019}$  \\
\rowcolor{lightgray}
AT (\bemph{I}) & $ 0.246_{.009}$ & $ \mathbf{0.233_{.005}}$ & $ 0.273_{.061}$ & $ 0.274_{.062}$  \\
\rowcolor{lightgray}
\phantom{AT} (\bemph{B}) & $ 0.140_{.004}$ & $ 0.128_{.005}$ & $ \mathbf{0.126_{.002}}$ & $ \mathbf{0.126_{.002}}$  \\
\phantom{MT} (\bemph{R}) & $ \mathbf{0.512_{.009}}$ & $ 0.521_{.008}$ & $ 0.535_{.010}$ & $ 0.539_{.009}$  \\
MT (\bemph{I}) & $ 0.396_{.019}$ & $ \mathbf{0.330_{.017}}$ & $ 0.402_{.036}$ & $ 0.413_{.038}$  \\
\phantom{MT} (\bemph{B}) & $ \mathbf{0.111_{.008}}$ & $ 0.133_{.007}$ & $ 0.135_{.006}$ & $ 0.134_{.006}$  \\
\rowcolor{lightgray}
\phantom{BQ} (\bemph{R}) & $ \mathbf{1.481_{.019}}$ & $ 1.499_{.019}$ & $ 1.533_{.019}$ & $ 1.537_{.018}$  \\
\rowcolor{lightgray}
BQ (\bemph{I}) & $ \mathbf{0.981_{.015}}$ & $ 0.994_{.014}$ & $ 1.033_{.016}$ & $ 1.041_{.016}$  \\
\rowcolor{lightgray}
\phantom{BQ} (\bemph{B}) & $ 0.099_{.009}$ & $ 0.095_{.004}$ & $ \mathbf{0.094_{.003}}$ & $ \mathbf{0.094_{.003}}$  \\
\phantom{RS} (\bemph{R}) & $ 0.598_{.006}$ & $ \mathbf{0.593_{.005}}$ & $ 0.608_{.005}$ & $ 0.603_{.005}$  \\
RS (\bemph{I}) & $ 0.537_{.023}$ & $ \mathbf{0.531_{.023}}$ & $ 0.553_{.022}$ & $ 0.549_{.024}$  \\
\phantom{RS} (\bemph{B}) & $ \mathbf{0.149_{.002}}$ & $ 0.297_{.003}$ & $ 0.327_{.006}$ & $ 0.326_{.006}$  \\
\rowcolor{lightgray}
\phantom{PE} (\bemph{R}) & $ \mathbf{1.006_{.013}}$ & $ 1.013_{.012}$ & $ 1.046_{.014}$ & $ 1.055_{.013}$  \\
\rowcolor{lightgray}
PE (\bemph{I}) & $ 0.587_{.128}$ & $ \mathbf{0.363_{.092}}$ & $ 0.755_{.129}$ & $ 0.797_{.127}$  \\
\rowcolor{lightgray}
\phantom{PE} (\bemph{B}) & $ 0.165_{.003}$ & $ \mathbf{0.142_{.003}}$ & $ 0.149_{.003}$ & $ 0.148_{.003}$  \\
\Xhline{1.5pt}
\end{tabular}
\end{table}
{\color{black}\subsection{Random Imputation Tasks (RQ5)}

By default, we evaluate \algo on three imputation granularities: \textit{R} (coarse-grained), \textit{I} (intermediate), and \textit{B} (fine-grained), as detailed in \hyperref[subsec:evalframework]{Sec.~5.1}. However, this does not limit its generalisability to other scenarios. To show this, we also compare the accuracies (MSE) on random imputation scenarios. These tasks are constructed by randomly masking out a certain percentage of signal rows  while keeping the metadata intact. Results are shown in \autoref{tab:randomimp}. As expected, increasing the proportion of missing information generally degrades performance. On comparing these results to the \textit{R} tasks (root level) in \autoref{tab:mse_table}, we find that MSE scores remain consistently better than the \textit{R} tasks, which have 100\% missing signals (hardest scenario). This highlights that \algo is not limited to specific missingness configurations and is generalisable to other scenarios.}

\begin{table}[htb]
    \centering
    \caption{{\color{black}MSE for varying missingness ratios}}
    \begin{tabular}{l lll}
        \bottomrule
        \Xhline{1.5pt}
        
        Dataset & 25\% &50\% & 75\%\\
        \midrule
        
MT & $ 0.092_{.003}$ & $ 0.104_{.002}$ & $ 0.137_{.003}$ \\

BAQ & $ 0.146_{.004}$ & $ 0.167_{.003}$ & $ 0.295_{.005}$ \\
RS & $ 0.233_{.007}$ & $ 0.271_{.008}$ & $ 0.351_{.008}$ \\
PE & $ 0.004_{.000}$ & $ 0.009_{.000}$ & $ 0.030_{.001}$ \\
AT & $ 0.141_{.003}$ & $ 0.140_{.003}$ & $ 0.138_{.005}$ \\

        \bottomrule
        \Xhline{1.5pt}
    \end{tabular}
    \label{tab:randomimp}
    \vspace{-5pt}
\end{table}
\subsection{Visual Assessment of Generated Data}
\label{ssec:visual}
\autoref{fig:quality} compares  \algo against other baselines: TSDiff and TimeWeaver, on the BQ dataset. \autoref{fig:acfbeijing} illustrates the autocorrelation for each task across 100 lags, along with the real signal.

We see that methods conditioning on observed signals (TSDiff and \algo) better resemble ground truth signals under finer-grained tasks. In contrast, TimeWeaver does significantly worse on bottom-level tasks since it does not condition on the observed signals. Cross-feature correlations (see \autoref{fig:crosscorrbeijing}) further highlight this trend, with \algo and TSDiff maintaining better inter-feature dependencies for fine-grained tasks. These results highlight \algo's strength in capturing key temporal and structural patterns.

\section{Related Work}
\autoref{tab:relwork} summarises the related works from time series generation  and adjacent domains. These works vary in their backbone architectures (GANs versus diffusion); their ability to condition on both metadata and observations; and their conditioning strategies (training versus inference-time). 

\textbf{Time Series Models:} Generative models have evolved in recent years. TimeGAN~\cite{yoon2019time} pioneered GAN-based architectures for time series generation, but focuses on unconditional synthesis without leveraging metadata or the observed signals. In contrast, more recent methods employ conditioning to utilise additional information for guiding the generative process.

TimeWeaver \cite{narasimhan2024time}  trains a conditional diffusion backbone by generating the entire signal just from the metadata. However, it ignores the observed signals, despite their criticality for fine-grained imputation tasks. In contrast, TimeGrad \cite{rasul21a} ignores metadata but only handles forecasting tasks by conditioning on historical signals from preceding timesteps, which is unsuitable for imputation at arbitrary timesteps. Moreover, it maintains temporal coherence via autoregression, which is slow due to sequential generation.
SSSD~\cite{alcaraz2022diffusion} includes the conditional masks as inputs during training. While this enables imputing values at arbitrary timesteps, it risks overfitting on the masking patterns seen during training. TSDiff~\cite{kollovieh2024predict} removes this risk by conditioning on the observed signal patterns directly at inference through self-guidance. However, self-guidance only conditions on the observations within a given time window, ignoring dependencies across overlaps. 
TimeAutoDiff~\cite{suh2024timeautodiff} sidesteps this cross-window dependency problem by generating the entire output in one go using a latent transformer-based diffusion model \cite{rombach2022high} and classifier-free guidance~\cite{ho2021classifier} for conditioning on labels. However, this design restricts it to fixed-length sequences and is impractical: longer sequences significantly increase the computational load on the self-attention layers, hurting scalability~\cite{keles2023computational}. Moreover, adapting classifier-free guidance from conditioning on discrete labels to conditioning on the continuous and arbitrarily observed signals is challenging.
\begin{figure*}
\centering
    \begin{subfigure}{0.45\textwidth}
        \centering
        \includegraphics[width=\textwidth]{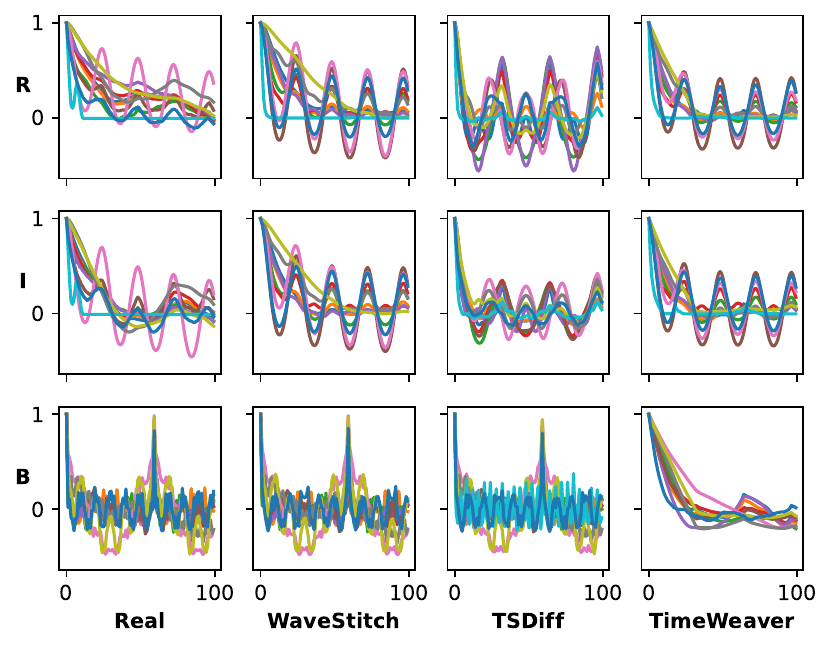}
        \caption{Per-channel Autocorrelation}
        \label{fig:acfbeijing}
    \end{subfigure}
    \begin{subfigure}{0.45\textwidth}
        \centering
        \includegraphics[width=\textwidth]{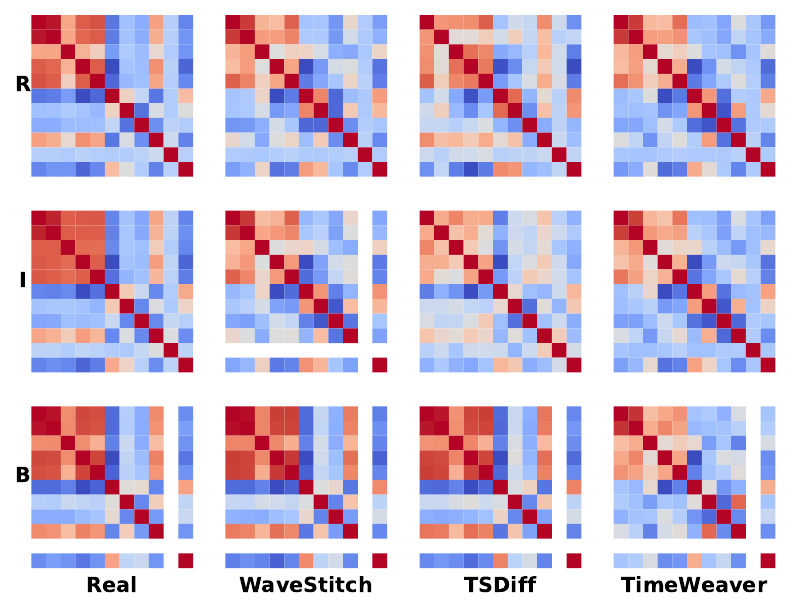}
        \caption{Cross-feature Correlations}
        \label{fig:crosscorrbeijing}
    \end{subfigure}
    
    \caption{Autocorrelation (up to 100 lags), and Cross-feature Correlations for BQ for \emph{R}, \emph{I}, and \emph{B} tasks.}
    \label{fig:quality}
\end{figure*}

\textbf{Non-Time-Series Models:}   
RePaint~\cite{lugmayr2022repaint}, designed for image inpainting, perturbs the observed pixels to blend their noise with that of the missing regions. Then, it jointly denoises both using an unconditional model directly at inference. However, independently noising the observed areas leads to noise misalignment with the missing regions, producing semantically inconsistent outputs~\cite{chung2022improving,lugmayr2022repaint}. For video generation, NUWA-XL~\cite{yin2023nuwa} uses a hierarchical "coarse-to-fine" diffusion approach using two models. It first generates coarse keyframes using prompts on an unconditional model, then fills the gaps using classifier-free guidance~\cite{ho2021classifier}using a conditional one. However, its conditional model specifically uses the first and last frames and cannot be conditioned on arbitrary frame positions. Moreover, the recursive refinement introduces sequential dependencies: each refinement stage depends on the previous one, leading to autoregressive-like behaviour.  

In summary, many existing related works rely on minimal or rigid conditioning, ignore temporal dependencies, or are inefficient. Instead, \algo offers a lightweight, inference-time solution that flexibly accommodates diverse conditioning signals and maintains temporal consistency without retraining or sacrificing parallelism.

\begin{table}[thb]
    \caption{Summary of related works}
    \footnotesize
    \centering
    \begin{tabular}{>{\raggedright\fontsize{9}{11}\selectfont}p{2.7cm}c c c c c}
    \Xhline{1.5pt}
    \rotatebox{45}{Method} & \rotatebox{45}{Time Series} & \rotatebox{45}{\makecell{Metadata}} & 
    \rotatebox{45}{\makecell{Observed\\ Values}} &
    \rotatebox{45}{\makecell{Cross-window\\ Coherence}} & \rotatebox{45}{\makecell{Inference-time\\Conditioning}}\\ 
    
    TimeWeaver~\cite{narasimhan2024time} & \tealcheck & \tealcheck & \redcross & \redcross & \redcross \\
    TimeGAN~\cite{yoon2019time} & \tealcheck & \redcross & \redcross & \redcross & \redcross \\
    TSDiff~\cite{kollovieh2024predict} & \tealcheck & \redcross & \tealcheck & \redcross & \tealcheck \\
    SSSD~\cite{alcaraz2022diffusion} & \tealcheck & \redcross & \tealcheck & \redcross & \redcross \\
    TimeGrad~\cite{rasul21a} & \tealcheck & \redcross & \tealcheck & \tealcheck & \redcross \\    TimeAutoDiff~\cite{suh2024timeautodiff} & \tealcheck & \tealcheck & \redcross & \redcross & \redcross\\
    
    RePaint~\cite{lugmayr2022repaint} & \redcross & \redcross & \tealcheck & \redcross & \tealcheck \\

    NUWA-XL~\cite{yin2023nuwa} & \redcross & \tealcheck & \tealcheck & \tealcheck & \redcross \\
    \midrule
    \algo (ours) & \tealcheck & \tealcheck & \tealcheck & \tealcheck & \tealcheck \\
    
    
    \Xhline{1.5pt}        
    \end{tabular}
    \vspace{-5mm}
    \label{tab:relwork}
\end{table}

\section{Conclusion}
We present \algo, a novel framework for conditional time series synthesis. \algo makes three key contributions: {\color{black}(1) integrating metadata and observed signals through a dual-sourced conditioning strategy; (2) hybridising training and inference via metadata-conditioned model training and gradient-based refinements on the observed signals using a conditional loss; and (3) coherent pipelined parallel generation through overlap alignment via a stitching mechanism.} Built on a diffusion backbone, \algo integrates these innovations to generate high-quality, temporally consistent time series across various tasks. Empirical results demonstrate that \algo handles complex conditioning patterns and outperforms SOTA baselines with a \textcolor{black}{\textbf{1.81x}} lower MSE on average, and a speedup of up to \textbf{166.48x} over autoregressive methods. While effective, \algo currently treats metadata as static and fully observed. An important direction for future work is to extend the framework to handle partially observed metadata, which introduces additional challenges such as synthesising discrete and continuous-valued features. Other promising avenues include conditioning strategies for black-box settings, exploring alternative modalities for conditioning, and scaling \algo to distributed synthesis settings. These advances would further enhance \algo’s flexibility across tasks and enable conditional generation under data privacy and resource constraints.


\section*{Acknowledgements}
This publication was supported by the Dutch Research Council (VI.Veni.222.439), the Smart Networks and Services Joint Undertaking (SNS JU) under the European Union’s Horizon Europe Research and Innovation programme (Grant Agreement No. 101192750), the Swiss National Science Foundation (Priv-GSyn project, 200021E\_229204), and the DEPMAT project (P20-22 / N21022) of the research programme Perspectief which is partly financed by the Dutch Research Council (NWO).

\bibliographystyle{ACM-Reference-Format}
\bibliography{ref}
\end{document}